%% file: main.tex
\documentclass[lettersize,journal]{IEEEtran}
\usepackage{amsmath,amsfonts}
\usepackage{amssymb}
\usepackage{array}
\usepackage{textcomp}
\usepackage{url}
\usepackage{verbatim}
\usepackage{graphicx}
\usepackage[caption=false,font=footnotesize,labelfont=sf,textfont=sf]{subfig}
\usepackage{stfloats}
\usepackage{cite}
\usepackage{bbding}
\usepackage{pifont}
\newcommand{\cmark}{\ding{51}}
\newcommand{\xmark}{\ding{55}}
\usepackage{color}
\usepackage{colortbl}
\definecolor{Gray}{gray}{0.9}
\usepackage{booktabs}
\usepackage{multirow}
\usepackage[ruled,lined]{algorithm2e}
\usepackage{enumitem}
\usepackage[pagebackref,breaklinks,colorlinks]{hyperref}

\newcommand{\etal}{\textit{et al}.}
\newcommand{\ie}{\textit{i}.\textit{e}.}
\newcommand{\eg}{\textit{e}.\textit{g}.}

\begin{document}
	
	\title{Co-Learning Meets Stitch-Up for Noisy Multi-label Visual Recognition}
	
	\author{Chao Liang, Zongxin Yang, Linchao Zhu, Yi Yang
		\thanks{Chao Liang, Zongxin Yang, Linchao Zhu, Yi Yang are with School of Computer Science, Zhejiang University, Zhejiang, China. E-mail: \{cs.chaoliang, yangzongxin, zhulinchao, yangyics\}@zju.edu.cn.
		This work is supported by National Key R\&D Program of China under Grant No. 2020AAA0108800. This work is partially supported by the Fundamental Research Funds for the Central Universities (No. 226-2022-00051)}
	}
	
	\markboth{Journal of \LaTeX\ Class Files,~Vol.~14, No.~8, August~2021}%
	{Shell \MakeLowercase{\textit{et al.}}: A Sample Article Using IEEEtran.cls for IEEE Journals}
	
	\IEEEpubid{}
	
	\maketitle
	
	\begin{abstract}
		In real-world scenarios, collected and annotated data often exhibit the characteristics of multiple classes and long-tailed distribution. Additionally, label noise is inevitable in large-scale annotations and hinders the applications of learning-based models. Although many deep learning based methods have been proposed for handling long-tailed multi-label recognition or label noise respectively, learning with noisy labels in long-tailed multi-label visual data has not been well-studied because of the complexity of long-tailed distribution entangled with multi-label correlation. To tackle such a critical yet thorny problem, this paper focuses on reducing noise based on some inherent properties of multi-label classification and long-tailed learning under noisy cases. In detail, we propose a Stitch-Up augmentation to synthesize a cleaner sample, which directly reduces multi-label noise by stitching up multiple noisy training samples. Equipped with Stitch-Up, a Heterogeneous Co-Learning framework is further designed to leverage the inconsistency between long-tailed and balanced distributions, yielding cleaner labels for more robust representation learning with noisy long-tailed data. To validate our method, we build two challenging benchmarks, named VOC-MLT-Noise and COCO-MLT-Noise, respectively. Extensive experiments are conducted to demonstrate the effectiveness of our proposed method. Compared to a variety of baselines, our method achieves superior results.
	\end{abstract}
	
	
	\section{Introduction}
	\IEEEPARstart{T}{he} remarkable breakthroughs of convolutional neural networks~\cite{he2016deep,krizhevsky2012imagenet,zhang2021weakly} in visual recognition can be largely attributed to the arising of large-scale data resources. In image recognition, conventional classification settings~\cite{deng2009imagenet,he2016deep,krizhevsky2012imagenet} typically assume that each image is annotated with a single-label and each class contains the same number of instances. In real-world scenarios, however, collected data often exhibit the characteristics of long-tailed distribution~\cite{zhou2020bbn,liu2019large,cao2019learning,kang2019decoupling} and multi-label annotations~\cite{wang2016cnn,chen2019multi, ji2020deep}. Besides, label noise~\cite{sun2017revisiting, wu2021bispl, ye2021collaborative, huang2021scribble} generally exists in large-scale annotations. Training models with noisy labels inevitably degenerates networks' learning performance~\cite{li2020dividemix,li2020mopro} and thus hinders the 
	development of robust learning-based models.
	
	\input{figures/figure1}
	
	For improving networks' classification performance under long-tailed multi-label scenarios, several recent works extend widely used long-tailed methods in the single-label setting to the multi-label case. Wu~\etal~\cite{wu2020distribution}~proposed a distribution balanced loss to deal with label co-occurrence. Guo~\etal~\cite{guo2021long}~leveraged the network consistency loss to learn a robust representation under different sampling strategies. As to handle with label noise, current methods~\cite{li2020dividemix,han2018co,li2020mopro,ren2018learning} on learning with noisy labels mostly focus on a single-label setting. DivideMix~\cite{li2020dividemix} divided the training data into the labeled and unlabeled set, and then employed the semi-supervised way to tackle this problem. MoPro~\cite{li2020mopro} resorted to label correction to mitigate the impact of noisy labels. Although the above progress is remarkable, the problem of label noise in long-tailed multi-label data has been barely explored or well-studied since the combination of long-tailed distribution and multi-label correlation will further complicate the label noise' problem. When encountered with long-tailed multi-label data, existing strategies neglect the properties of multi-label annotations~\cite{ren2018learning} or fail to generalize well under imbalanced data distribution~\cite{li2020dividemix,li2020mopro}. 
	
	In order to relieve the noise problem within the challenging long-tailed multi-label scenarios, we go back to one of the keys of learning with noisy labels, \ie, reducing noise in the training stage, based on some inherent properties of multi-label classification and long-tailed learning under noisy cases. \textbf{First}, the negative effect of noisy labels can be alleviated when training with cleaner data. Multi-label classification aims to detect the existence of the object in a given image for each class. In noisy cases, a set of labeled images (with the same label) is more likely to contain all labeled classes than any single image from the set. As shown in Figure~\ref{fig:intro}, when given only one image with a noisy multi-label containing the cat, the model is uncertain to predict whether the cat exists in this image or not. However, when a pair of images is provided to tell the cat's presence, the probability of the cat's existence highly increases. In other words, if we stitch up a set of images and take the union of their labels, a training example with less noisy possibility can be synthesized. \textbf{Second}, disagreement from different sampling priors helps us distinguish noisy labels when the training distribution is long-tailed. Label correction is an effective tool to correct noisy labels in the literature~\cite{li2020mopro,yang2020webly}.
	Differently, Co-teaching\cite{han2018co} leveraged the inconsistency between two networks to select clean samples. This inconsistency is obtained from the different network initialization. Motivated by these methods, our intuition is that the disagreement~\cite{yu2019does} under different sampling policies can help us correct noisy labels in the distribution of long-tailed. More specifically, random sampling prefers head classes, while class re-balanced sampling tends to handle the tail better. We inherit such a two-branch structure to rectify noisy labels from cross supervision using the discrepancy between different sampling priors.
	
	Based on the motivations mentioned above, this paper tackles the multi-label long-tailed classification problem with noisy labels. First, we propose a simple but effective augmentation called ``\textit{Stitch-Up}" to synthesize cleaner training samples directly. Specifically, Stitch-Up concatenates several images sharing the same classes and unions their labels simultaneously. Such a strategy can reduce label noise in the training data from the point of the probability theory. On the other hand, it can preserve information lossless.
	Second, a \textit{Heterogeneous Co-Learning} framework is designed to perform label correction by exploiting the sampling prior. Our framework consists of two branches jointly trained with random and balanced sampling. Each branch rectifies noisy labels for those with high confidence and these corrected pseudo-labels are used to guide the training procedure of its peer network. The above two branches can recognize different noisy samples, which benefits from the heterogeneous structure with different sampling distributions. And Co-Learning can reduce error accumulation caused by training with wrong labels so that the performance is improved. 
	
	We conduct extensive experiments on the noisy version of two multi-label long-tailed datasets named VOC-MLT-Noise and COCO-MLT-Noise, respectively. Compared to a variety of baseline methods, our superior results demonstrate the effectiveness of the proposed method.

	The main contributions are summarized as follows:
	\begin{itemize}
		\item We introduce a novel \textit{Stitch-Up} augmentation to synthesize cleaner training samples. The generated cleaner training data facilitate the learning of more robust models.
		\item By leveraging different sampling priors and loss functions, we design a \textit{Heterogeneous Co-Learning} framework to rectify noisy labels. The corrected pseudo-labels from one network cross-guide the training process of its peer network, which boosts the network's performance.
		\item To validate the effectiveness of our method, we propose two synthetic multi-label long-tailed benchmarks named VOC-MLT-Noise and COCO-MLT-Noise, respectively, with multiple noisy rates. 
		\item Thorough experiments on these two datasets demonstrate the effectiveness of our method.
	\end{itemize}
	
	\section{Related Work}
	\subsection{Long-tailed recognition}
	When learning with long-tailed data, one of the obstacles is that the frequent classes dominate the training procedure. But the test criterion typically prefers a uniform distribution or places more attention on the less representative classes~\cite{cao2019learning}. Such inconsistency between the training and test phase leads to poor performance.

	\input{figures/figure2}

	A wide range of strategies have been proposed to mitigate the effect of long-tailed distribution, including data resampling~\cite{shen2016relay,mahajan2018exploring,zhou2020bbn, he2009learning}, cost-sensitive reweighting~\cite{cui2019class,khan2017cost,tan2020equalization, wang2017learning}, margin aware loss~\cite{li2019overfitting,cao2019learning}, two-stage finetune~\cite{cui2018large,kang2019decoupling}, transfer learning~\cite{zhong2019unequal,liu2019large} and meta-learning~\cite{liu2019large,jamal2020rethinking}. Among them, resampling and reweighting are the two most prominent directions. Resampling methods tend to adjust a more balanced training distribution.~\cite{shen2016relay} advocates oversampling the minority classes while~\cite{he2009learning} claims that undersampling the frequent classes is better. In recent work, Zhou~\etal~\cite{zhou2020bbn} proposed a two-branch architecture called BBN with different sampling strategies. The unified conventional and re-balancing branches promote both representation and classifier learning. Reweighting techniques~\cite{huang2019deep,wang2017learning} assign large weights for the training samples in the tail classes, to resist the skew prior distribution.

	The majority of the aforementioned works focus on the multi-class setting, where each image has a single label. In real-world scenarios, large-scale datasets are often annotated with multiple labels and exhibit a long-tailed distribution~\cite{sun2017revisiting}. Recently, Wu~\etal~\cite{wu2020distribution} proposed to combine rebalancing and reweighting methods to handle long-tailed multi-label recognition. Guo~\etal~\cite{guo2021long} extended the BBN~\cite{zhou2020bbn} framework to support multi-label learning, enforcing consistency between different branches. Despite their improved performance, they ignored the label noise and assumed the collected datasets are clean. Our work targets handling the label noise in the multi-label long-tailed setting.
	
	\subsection{Multi-label classification}
	Before the era of deep learning, multi-label classification is primarily tackled by turning it into multiple independent binary classification problems~\cite{tsoumakas2007multi} or adapting the existing algorithms, e.g. k-nearest neighbors~\cite{zhang2007ml}, decision tree~\cite{clare2001knowledge}, kernel learning~\cite{elisseeff2001kernel}. However, training separate binary classifiers neglects the relationships between labels and it is impractical to enumerate all the label combinations. As ConvNets receive significant success, modern methods~\cite{wang2016cnn,lee2018multi, chen2019multi,wang2021semi, xie2017sde, sun2021compositional} rely on deep networks to model the label dependencies. In~\cite{wang2016cnn}, the recurrent neural network is utilized to embed the label correlations. Lee~\etal~\cite{lee2018multi} leveraged the knowledge graph~\cite{yang2021multiple} to describe the relationships between labels. ML-GCN~\cite{chen2019multi} captured the label co-occurrence by the graph structures. SDE~\cite{xie2017sde} sought for a selective, discriminative and equalizing feature representation by a learning-based feature pooling framework.
	
	Contrary to their works, we focus on the multi-label noise in the image classification. Since the images annotated with the multi-label often share the same label, the label noise can be reduced in the training set when we synthesize cleaner images by stitching up collections of such images.
	
	\subsection{Learning with noisy labels} Deep neural networks are prone to fit noisy labels~\cite{zhang2021understanding}. Training with corrupted labels can inevitably yield poor generalization performance. Existing works on learning with noisy labels can be roughly divided into three categories: (1) sample selection~\cite{li2020dividemix,han2018co,yu2019does,arazo2019unsupervised}. It works by filtering out label noise and retraining with clean data. Small loss trick plays an important role in noise identification, based on the observation~\cite{arpit2017closer} that deep neural networks often memorize simple patterns first and then noisy samples. (2) label correction~\cite{li2020mopro,yang2020webly}. Unreliable supervision from noisy labels can make optimization difficult. Several methods perform label correction by prediction from the network.  (3) sample reweighting~\cite{ren2018learning,shu2019meta, xu2021training}. This approach is a commonly used strategy against noisy labels. Ren \etal~\cite{ren2018learning} allocated weights based on the gradient direction. Meta-Weight-Net~\cite{shu2019meta} adopted the meta-learning framework to learn a weighting function mapping from the training loss. 
	
	These methods mostly address single-label noise and have great limitations in the long-tailed and multi-label scenarios. Our approach considers the properties of multi-label and long-tailed distribution. We propose a heterogeneous structure that allows better label correction during the training procedure.
	
	\section{Method}
	\subsection{Overview}
	Our proposed method aims to address the multi-label long-tailed classification problem with noisy labels. Suppose that we are given a training dataset $\mathcal{D}_{train}=\{(\mathbf{x}_i, \mathbf{\tilde{y}}_i)|i=1,2,...,N\}$, where $N$ is the number of the training samples and each sample is annotated with a noisy multi-label $\mathbf{\tilde{y}}_i$. Specifically, $\mathbf{\tilde{y}}_i \in \{0,1\}^C$ contains $C$ binary labels with 1 indicating the presence of the label and 0 otherwise. And $\mathbf{\tilde{y}}_i$ might be incomplete or mislabeled with more absent categories compared to clean label $\mathbf{y}_i$. In addition, the number of samples per class is imbalanced. The goal of this task is to learn a robust model with well generalization ability on the unseen test data when training on a multi-label long-tailed noisy dataset.
	
	In this work, we expect to handle label noise under the multi-label long-tailed setting. As deep models are prone to memorizing wrong labels~\cite{arpit2017closer}, learning with noisy labels poses great challenges to train deep neural networks effectively. One of the keys to alleviating the negative effect of noisy labels is reducing noisy training samples. Following this direction, we take the advantage of inherent properties within multi-label long-tailed circumstances to combat label noise. First, we propose a novel Stitch-Up augmentation to obtain less noisy training samples (Section~\ref{sec:stitchup}). We stitch up multiple images and their multi-labels simultaneously. This results in cleaner training samples. Second, we introduce a Heterogeneous Co-Learning framework to perform online noisy label correction in  Section~\ref{sec:colearning}. By leveraging inconsistency between different sampling priors, we rectify the wrong labels based on the confidence from our model. The corrected pseudo-labels are utilized to cross-guide the training procedure of its peer network. Then, we introduce the overall pipeline in Section~\ref{sec:pipeline}. In the end, Section~\ref{sec:loss} and Section~\ref{sec:inference} details the loss function we optimize and inference procedure. 
	
	\subsection{Stitch-Up}
	\label{sec:stitchup}
	Intuitively, training with less noisy examples can boost the model's performance. Since multi-label visual recognition aims to predict the existence of the object in an image for each class, it is more likely to find the presence of a class when given a set of images. This motivates us to synthesize new cleaner training samples by stitching up a collection of images that may share the same label in noisy cases.
	
	For each training sample $\mathbf{x}_i$ with a multi-label $\mathbf{\tilde{y}}_i$, we stitch up a set of examples with overlapping labels in the training set. \textbf{Sample Selection:} We construct the candidate set $\mathcal{S}_i^k$ composed of $K$ samples that share the same class $k$ for Stitch-Up. Specifically, in the first step, we choose an existing object class $k$ from $\mathbf{\tilde{y}}_i$, where $\tilde{y}_{ik} = 1$. In the second step, a collection of $K-1$ samples with class $k$ are selected from the subset of the training data $\mathcal{D}_{train}^k=\{(\mathbf{x}_j, \mathbf{\tilde{y}}_j)| \tilde{y}_{jk}=1\}$. Combined with the original sample $(\mathbf{x}_i, \mathbf{\tilde{y}}_i)$, these $K$ samples are formed as $\mathcal{S}_i^k$ where $|\mathcal{S}_i^k| = K$. \textbf{Stitch-Up Synthesis:} Then, we obtain a new training sample $(\mathbf{\bar{x}}_i, \mathbf{\bar{y}}_i)$ by stitching up these $K$ samples and performing label union. This process can be expressed as:
	\begin{align}
		\mathbf{\bar{x}}_i = \bigcup_{(\mathbf{x}_j, \mathbf{\tilde{y}}_j) \in \mathcal{S}_i^k}{\mathbf{x}_j}, \\
		\mathbf{\bar{y}}_i = \bigcup_{(\mathbf{x}_j, \mathbf{\tilde{y}}_j) \in \mathcal{S}_i^k}{\mathbf{\tilde{y}}_j}.
	\end{align}
	
	When training with deep neural networks, we treat Stitch-Up as means of data augmentation and apply this augmentation with a probability of $p$. Note that our Stitch-Up can be also applied at the feature level. In practice, Stitch-Up can be implemented in various forms. We show three regular types of Stitch-Up augmentation in Figure~\ref{fig:framework_stitchup}. All three types perform label union as means of Label Stitch-Up. For input images concatenation, we concatenate the images directly after sample selection and then feed the concatenated image into deep models. For features concatenation, we obtain the intermediate feature for each image and then concatenate these features. For features average, we average the intermediate features instead. The intermediate features can be extracted from different stages of deep neural networks. Empirically, we adopt the features average in the experiments.
	
	\textbf{Why does Stitch-Up work?}
	We provide a simple explanation based on the probability theory. For an object class $k$ with a noise rate of $\gamma_k$~($0<\gamma_k<1$), the probability of the existence for class $k$ is $1-\gamma_k$ when we are given only one image annotated with a noisy multi-label containing class $k$. If we stitch up two such images that share the same class $k$, we get a higher probability of $1-\gamma_k^2$. This reveals that stitch-up augmentation can reduce label noise explicitly.
	
	\textbf{Comparison against Mix-Up.} Our Stitch-Up is similar to Mix-Up~\cite{zhang2017mixup} augmentation. Both combine the samples and labels simultaneously to synthesize new training samples. However, the motivation is quite different. Mix-Up encourages the model to behave linearly, reflecting a good inductive bias, while Stitch-Up can synthesize more training samples with less label noise. Besides, our Stitch-Up can benefit from lossless information. Mix-Up linearly interpolates two images, which suffers from the unnatural artifact problem~\cite{yun2019cutmix}. The linear interpolation result looks unnatural and can discard the information to some extent. In contrast, our Stitch-Up concatenates the input images. This can leave the image intact without losing information. Our experiment further verifies Stitch-Up can outperform Mix-Up under the multi-label long-tailed recognition with noisy labels. The information loss might affect the head, medium and tail classes differently due to multi-label noise.
	
	\subsection{Heterogeneous Co-Learning}
	\label{sec:colearning}
	In this section, we introduce a Heterogeneous Co-Learning framework to overcome the overfitting when training with noisy labels. As Co-Learning\cite{han2018co,li2020dividemix} shows promising results in dealing with single-label noise, the main idea is to leverage the inconsistency through different network initialization to select clean training examples. In the long-tailed situation, we notice that different sampling priors have different preferences. To be more specific, classifiers trained with random sampling tend to behave well on head classes while those under balanced sampling can recognize tail classes better. Based on this observation, we propose to exploit the sampling prior to detecting noisy labels. We enforce the network to be cross-guided by the pseudo-labels which corrected by its peer network.
	
	As illustrated in Figure~\ref{fig:framework_colearning}, we jointly train two branches $f$ and $g$ with different sampling strategies. The first branch $f$ takes the uniform sampling distribution, where each instance has the same sampling probability $\frac{1}{N}$. The second branch $g$ adopts the class-rebalanced sampling and each class achieves an equal probability $\frac{1}{C}$ of being selected. Both $f$ and $g$ are shared with the same backbone $\Phi$.
	
	We rely on the output from the network to rectify noisy labels.
	In the multi-label classification problem, each label only has two states: 1 indicates the existence of the class and 0 otherwise. Therefore, we can perform label correction separately. \textbf{Pseudo Labeling:} We convert the probability $q_{ik}$ produced by the network $f$(or $g$) into pseudo-label $\hat{y}_{ik}$ (Eq.~\ref{eq:pseudo_label}) based on the following rules: if the probability $q_{ik}$ is extremely high or low, which is above or below some certain threshold~($\alpha$ or $\beta$), we trust the confidence from the network. Otherwise, we keep the original noisy label unchanged. We define the whole process as follows:
	\begin{equation}
		\hat{y}_{ik} = \begin{cases}1, & \text{if}~ q_{ik} > \alpha, \\ 
			0, & \text{if}~ q_{ik} < \beta,\\
			\tilde{y}_{ik}, & \text{otherwise}.\end{cases}
		\label{eq:pseudo_label}
	\end{equation}
	
	When noisy labels are corrected, we use the generated pseudo-labels to teach the learning of the other branch. Here, we take the random sampling branch as an example. For the training example $(\mathbf{x}, \mathbf{\tilde{y}})$ sampled from uniform distribution, we feed it into two branches. Then, given the output $g(\Phi(\mathbf{x}))$ from the class-rebalanced branch and the noisy label $\mathbf{\tilde{y}}$, we perform label correction and obtain the pseudo-label $\mathbf{\hat{y}}$. The pseudo-label is subsequently used to directly guide the training procedure of the random branch $f$. 
	
	Co-Learning benefits from the heterogeneous structure between long-tailed and balanced distributions. The inconsistency helps label correction, which could potentially improve the robustness of the learned model.
	
	\textbf{Relations to Co-Learning based approaches.} We compare our proposed method with other Co-Learning based approaches. Although our method is motivated by other Co-Learning idea~\cite{han2018co, li2020dividemix}, there are fundamental differences. Our \textit{Heterogeneous} Co-Learning is designed to combat label noise in long-tailed multi-label data. First, to tackle the long-tailed distribution problem, we leverage different sampling priors and loss functions for two branches where random sampling prefers head classes and class re-balanced sampling tends to handle the tail better. In contrast, other Co-Learning based approaches~\cite{han2018co,li2020dividemix} use different network initialization to help filter out label noise without the consideration of the long-tailed issue. They keep the sampling strategy and loss function the same for the two branches. Second, we exploit the predictions from the network to correct noisy multi-labels directly. Instead, other Co-Learning based approaches~\cite{han2018co,li2020dividemix} use the small-loss criterion to select clean data. The selected clean data are utilized to train the network. These approaches are designed for single-label noise. When encountered with multi-label noise, some data might have partially correct labels. It is hard to select totally clean data. Label correction is a more effective way to handle multi-label noise.
	
	\input{algo/algorithm}
	
	\subsection{Overall framework}
	\label{sec:pipeline}
	We illustrate our Heterogeneous Co-Learning framework equipped with Stitch-Up. We take $K = 2$ for example. The full algorithm is described in Algorithm~\ref{alg:overall}. The overall framework is presented in Figure~\ref{fig:framework}. Given two subsets $(\mathbf{X}_1, \mathbf{\widetilde{Y}}_1)$ and $(\mathbf{X}_1', \mathbf{\widetilde{Y}}_1')$ by random sampling and class re-balanced sampling from $\mathcal{D}_{train}$ respectively, we perform \textbf{Sample Selection}~(Section~\ref{sec:stitchup}) for each training sample in $\mathbf{X}_1$ and $\mathbf{X}_1'$ to obtain $(\mathbf{X}_2, \mathbf{\widetilde{Y}}_2)$ and $(\mathbf{X}_2', \mathbf{\widetilde{Y}}_2')$ for Stitch-Up. For each $(\mathbf{x}_1, \mathbf{x}_2) \in (\mathbf{X}_1, \mathbf{X}_2)$ with the noisy multi-label $(\mathbf{\tilde{y}}_1, \mathbf{\tilde{y}}_2) \in (\mathbf{\widetilde{Y}}_1, \mathbf{\widetilde{Y}}_2)$, we feed them into random branch and take \textbf{Stitch-Up Synthesis}~(Section~\ref{sec:stitchup}) at the feature level:
	\begin{align}
	\label{eq:stitch_feature}
		\mathbf{\bar{f}} = (f_1(\Phi(\mathbf{x}_1)) + f_1(\Phi(\mathbf{x}_2)))/2.
	\end{align}
	Then, logit $\mathbf{\bar{z}}$ is generated from $f_2$:
	\begin{align}
	\label{eq:gen_logit}
		\mathbf{\bar{z}} = f_2(\mathbf{\bar{f}}) .
	\end{align} 
	
	On the other hand, we cross input $(\mathbf{x}_1, \mathbf{x}_2)$ into the class-rebalanced branch to generate the label. First, we get two probabilities $\mathbf{q}_1$ and $\mathbf{q}_2$ from the branch $g$ as follows:
	\begin{align}
	\label{eq:q1}
		\mathbf{q}_1 = \boldsymbol{\sigma}(g_2(g_1(\Phi(\mathbf{x}_1)))), \\
	\label{eq:q2}
		\mathbf{q}_2 = \boldsymbol{\sigma}(g_2(g_1(\Phi(\mathbf{x}_2)))),
	\end{align}
	where $\boldsymbol{\sigma}$ denotes the sigmoid activation function. Second, each probability $\mathbf{q}_1$ and $\mathbf{q}_2$ is used to correct the corresponding original noisy label $\mathbf{\tilde{y}}_1$ and $\mathbf{\tilde{y}}_2$ via \textbf{Pseudo Labeling}~(see Eq.~\ref{eq:pseudo_label}). As a result, we obtain the pseudo labels $\mathbf{\hat{y}}_1$ and $\mathbf{\hat{y}}_2$. We produce the synthesized training label $\mathbf{\bar{y}}$ by Label Stitch-Up:
	\begin{align}
	\label{eq:lb_stitchup}
		\mathbf{\bar{y}} = \mathbf{\hat{y}}_1 \cup \mathbf{\hat{y}}_2.
	\end{align}
	
	In the end, the logit and new synthesized training label are fed into the loss function for the optimization of the network. For the class-rebalanced branch, we take the similar operation and get the logit $\mathbf{\bar{z}}'$ and the label $\mathbf{\bar{y}}'$.
	
	\subsection{Loss function}
	\label{sec:loss}
	The common approach to multi-label problems is to use the binary cross-entropy~(BCE) loss, which casts the multi-label classification as several binary classifications. We define the BCE loss as follows:
	\begin{align}
	\nonumber
	\label{eq:bce}
		\mathcal{L}_{BCE}(\mathbf{z}_i, \mathbf{y}_i) &=  -\frac{1}{C} \sum_{k=1}^{C} (y_{ik} \log(\boldsymbol{\sigma}(z_{ik})) ~+ \\
		&~~~(1-y_{ik}) \log(1-\boldsymbol{\sigma}(z_{ik}))),
	\end{align}
	where $\mathbf{z}_i$ denotes the logit, $\mathbf{y}_i$ is the multi-label and $\boldsymbol{\sigma}$ is the sigmoid function.
	However, the vanilla BCE loss fails to work when the training dataset also exhibits the long-tailed distribution~\cite{wu2020distribution}. Considering the label co-occurrence and negative classes dominance issues in the multi-label long-tailed recognition, Wu~\etal~\cite{wu2020distribution}~proposed Distribution-Balanced~(DB) loss under class-rebalanced sampling. Given the logit $\mathbf{z}_i$ and the multi-label $\mathbf{y}_i$, this loss is formulated as:
	\begin{align}
		\nonumber
		\label{eq:db}
		\mathcal{L}_{D B}(\mathbf{z}_{i}, \mathbf{y}_{i})&=-\frac{1}{C} \sum_{k=1}^{C} \hat{r}_{ik} (y_{ik} \log(\boldsymbol{\sigma}(z_{ik}-\nu_{k}))+\\
		&~~\frac{1}{\lambda}(1-y_{ik}) \log(1-\boldsymbol{\sigma}(\lambda(z_{ik}-\nu_{k})))),
	\end{align}
	where
	\begin{align}
		\hat{r}_{ik} &= \theta + \frac{1}{1+\exp(-\phi\times(r_{ik}-\mu))},\\
		r_{ik} &= \frac{\frac{1}{N_k}}{\sum_{y_{ij}=1}{\frac{1}{N_j}}},\\
		\nu_i &= \kappa\log(\frac{1}{p_k}-1).
	\end{align}
	Herein, $N_k$ denotes the number of instances in the class $k$, $r_{ik}$ is the re-balancing weight. $\nu_{k}$ represents the class-specific bias, $p_k = N_k/N$ is the class prior, $\boldsymbol{\sigma}$ is the sigmoid function and $\lambda, \theta, \phi, \mu, \kappa$ are hyper-parameters.
	
	Our two branches $f$ and $g$ are optimized by the different loss functions. For random branch $f$, we use the BCE loss while DB with Focal Loss\cite{lin2017focal} is applied to the class re-balanced branch $g$. The overall loss is summarized as:
	\begin{align}
		\mathcal{L} = \frac{1}{|\mathbf{X}_1|}\sum_{i=1}^{|\mathbf{X}_1|}\mathcal{L}_{BCE}(\mathbf{\bar{z}}_i, \mathbf{\bar{y}}_i) + \frac{1}{|\mathbf{X}_1'|}\sum_{i=1}^{|\mathbf{X}_1'|}\mathcal{L}_{DB-Focal}(\mathbf{\bar{z}}_i', \mathbf{\bar{y}}_i').
	\end{align}
	
	This inconsistency also prevents Co-Learning from degenerating to Self-Training and helps noisy label correction.
	
	\subsection{Inference}
	\label{sec:inference}
	To evaluate the test data, we ensemble the outputs from two branches. For an unseen image $\mathbf{x}$, we obtain the output $\mathbf{z}$:
	\begin{align}
		\mathbf{z} = \tau f(\Phi(\mathbf{x})) + (1-\tau) g(\Phi(\mathbf{x})),
	\end{align}
	where $\tau$ is a balanced factor.
	
	\section{Experiments}
	\subsection{Datasets}

	We evaluate the effectiveness of our method on two synthetic benchmark datasets: VOC-MLT-Noise and COCO-MLT-Noise. These datasets are artificially derived from VOC-MLT and COCO-MLT~\cite{wu2020distribution}, which are proposed for the evaluation of multi-label long-tailed image recognition. Since the original datasets are clean, we need to synthesize noisy labels. The details about noisy labels generation and the corrupted datasets are introduced as follows.
	
	\subsubsection{Noisy labels generation}
	Referring to the conventional ways to generate label noise in the single-label settings~\cite{han2018co,li2020dividemix}, we flip the  original clean labels by using the noise transition matrix. In the context of the multi-label long-tailed problem, we take the label co-occurrence and imbalanced distribution into consideration. We define the noise transition matrix $T_{ij}$ as the probability of being flipped to noisy label $j$ when given an instance with a clean label $i$. Formally, assume that noise rate $\gamma \in [0, 1]$, the noise transition matrix can be expressed as:
	\begin{align}
				\nonumber
				T_{i j}(X)&=\mathbb{P}(\bar{Y}=j \mid Y=i, X=x), \nonumber\\
				&=\begin{cases}1-\gamma, & j=i, \\ 
					\frac{N_{ij}}{\sum_{k\ne i}{N_{ik}}} \gamma, & j \ne i,\\\end{cases}
	\end{align}
	where $X$ is the training sample, $Y$ and $\bar{Y}$ represent the original clean label and the generated noisy label, respectively. Here, $N_{ij}$ denotes the number of instances in  frequency that the label $i$ and label $j$ co-occur in the dataset. Note that our construction does not care about the label combinations that can rarely appear in the same image, e.g. airplane and cow.
	
	In the experiments, we investigate the robustness of our method under the noise rate $\gamma \in \{0.3, 0.5, 0.7, 0.9\}$.
	
	\subsubsection{VOC-MLT-Noise}
	This dataset is extended from VOC-MLT dataset~\cite{wu2020distribution} by noisy labels generation. The original clean long-tailed dataset is sampled from VOC 2012 train-val set by pareto distribution. The training dataset consists of 1,142 images and 20 classes with a range from 4 to 775 images. Note that the label distribution can be slightly shifted after introducing label noise. We perform the evaluation on VOC2007 clean test set with 4,952 images.
	
	\subsubsection{COCO-MLT-Noise} COCO-MLT-Noise is constructed from COCO-MLT~\cite{wu2020distribution} in a similar way. This dataset is based on MS COCO-2017. There are 4,783 images from 80 classes in the training set.
	The maximum training samples per class of the original dataset is 1,356 and the minimum is 6. The test set is from COCO-2017 with 5,000 clean images.
	\input{tables/table_total}
	\input{tables/table_voc}
	
	\subsection{Implementation Details}
	\subsubsection{Training details}
	In our experiments, we use ResNet-50 pretrained on ImageNet as the backbone. The input images are randomly cropped and resized to $224\times 224$ with standard augmentation. The batch size is 32 for random sampling branch and 256 for class re-balanced sampling branch. We use SGD with momentum of 0.9 and weight decay of 0.0001 for optimization. We use linear warm-up for the first 100 iterations with a ratio of $\frac{1}{3}$. The total training epochs are 8 and the initial learning rate is cross-validated in \{0.02, 0.08, 0.14, 0.2\}, which decays by a factor 10 after 5 and 7 epochs. We follow the same DB-Focal loss configuration as \cite{wu2020distribution}. We use $K=2$ images for Stitch-Up augmentation with the probability of $p=1.0$. The hyperparameters $\alpha$ and $\beta$ is cross-validated in \{0.7, 0.8, 0.9\} and \{0.1, 0.2, 0.3, 0.4\}, respectively. The balanced factor $\tau$ for evaluation is 0.1.
	
	\subsubsection{Evaluation metric}
	Following~\cite{wu2020distribution,guo2021long}, we adopt the mean average precision~(mAP) to measure the performance. We report average mAP and the 95\% confidence interval over 5-trials for all classes and three subsets including \textit{head}, \textit{medium} and \textit{tail} classes. Head classes have more than 100 samples, medium classes contain 20-100 samples and those less than 20 samples are classified as tail classes. Besides, we also show mAP for each branches to observe the impact of our method.
	
	\subsubsection{Baseline settings}
	We compare our method with several baselines: (1) Empirical Risk Minimization~(ERM): This approach treats all the training instances with the same sampling probabilities and the same weights. We use the random sampling strategy and the BCE loss in the experiment. (2) Focal Loss~\cite{lin2017focal}: This loss is proposed to solve the class-imbalance problem. We set both the focusing parameter and the weighting factor to 2. (3) Re-Sampling~(RS)~\cite{shen2016relay}: We apply the class-rebalanced sampling with the vanilla BCE loss. (4) RS-Focal~\cite{shen2016relay}: This is the combination of the class-rebalanced sampling strategy and focal loss.  (5) Label Distribution Aware Margin loss~(LDAM)~\cite{cao2019learning}: This margin-based loss encourages each class has the optimal margin. We extend the original softmax-based implementation to BCE-based one for multi-label classification. (6) Bilateral-Branch Network~(BBN)~\cite{zhou2020bbn}: Similar to our method, this framework also inherits the two-branch structure with uniform and reversed samplers. This method considers the single-label long-tailed case without label noise. We make some modifications so that it can be adapted in the multi-label setting. (7) ML-GCN~\cite{chen2019multi}: A graph-based framework for multi-label image classification. (8) DivideMix~\cite{li2020dividemix}: A two-branch framework combines the sophisticated semi-supervised technique and sample selection to deal with single-label noise. We replace the sampling strategy and loss function for long-tailed and multi-label classification. Balanced sampling and BCE loss are used. (9) Distribution-Balanced loss~(DB)~\cite{wu2020distribution}: A recently proposed loss to solve multi-label classification in long-tailed datasets. (10) DB-Focal~\cite{wu2020distribution}: Compared to DB~\cite{wu2020distribution}, Focal Loss~\cite{lin2017focal} is further applied.
	
	\subsection{Results}
	Baseline methods are mostly based on one branch with random sampling or class-rebalanced sampling except BBN~\cite{zhou2020bbn} and DivideMix~\cite{li2020dividemix}. To verify the effectiveness of our proposed method, we conduct extensive experiments on our two synthetic benchmarks, VOC-MLT-Noise and COCO-MLT-Noise respectively, under the noise rate $\gamma \in \{0.3, 0.5, 0.7, 0.9\}$. We report the total mAP in Table~\ref{table:total}.
	
	First of all, we observe that with more training samples containing noisy labels, the performance is worse for all of the methods in both datasets. For VOC-MLT-Noise, the best total mAP under the noise rate $\gamma$ of 0.3 and 0.9 is 76.48\% and 34.41\%, respectively. The relative gap is around 42\%. For COCO-MLT-Noise, the best total mAP under the noise rate $\gamma$ of 0.3 is 21\% better than that in the noise rate $\gamma$ of 0.9. These results indicate that label noise can significantly hinder the learning of robust models in the multi-label and long-tailed setting. Second, DB and DB-Focal~\cite{wu2020distribution} are better than other baseline approaches since they consider both multi-label and long-tailed distribution. Especially when the noise rate is low~(\eg ~$\gamma = 0.3$), they still show much robust performance~(73.75\% and 72.87\%). Third, compared to several baseline methods, our method can gain significant improvements on mAP by reducing label noise. The total mAP of our proposed method on VOC-MLT-Noise in the different noise rates $\gamma \in \{0.3, 0.5, 0.7, 0.9\}$ is 76.48\%, 69.10\%, 62.29\%, 34.41\%, respectively. And the performance gap relative to the state of the art baseline is approximately \textbf{+2.7\%}, \textbf{+5.5\%}, \textbf{+7.2\%}, \textbf{+5.3\%}. For COCO-MLT-Noise, the improvement is about \textbf{+1.7\%}, \textbf{+2.0\%}, \textbf{+2.8\%} and \textbf{+3.2\%}, respectively. The lower confidence intervals also suggest that our method is more stable. When we take a closer look at the independent evaluation results of two branches, we find the class-rebalanced branch plays a more important role in the recognition. Furthermore, as seen in Table~\ref{table:VOC}, the mAP results on VOC-MLT-Noise in three subsets are presented. We find the performance is all improved for head, medium and tail. The recognition ability on the tail class is notably enhanced for the class-rebalanced branch. The class-rebalanced branch outperforms DB~\cite{wu2020distribution} by 3.2\% in the noise rate $\gamma$ of 0.5. In the meantime, random branch achieves consistent better performance for the head class. When the noise rate $\gamma$ is 0.5, our random branch gets 65.39\% mAP for the head classes, which is 3.4\% better than BBN~\cite{zhou2020bbn}. Compared to DivideMix~\cite{li2020dividemix}, our proposed method achieves about 9.8\% improvement on VOC-MLT-Noise under the noise rate of 0.5. Although Dividemix is designed to deal with label noise, it does not consider the long-tailed and multi-label issues. Table~\ref{table:COCO} shows the head, medium and tail performance on COCO-MLT-Noise. We notice that our method achieves 56.78\%~(+3.2\%) mAP for the head classes and 45.17\%~(+1.7\%) mAP for the tail classes. These results confirm the superiority of our proposed method.
	\input{tables/table_coco}
	
	\input{tables/table_ablation}
	
	\subsection{Ablation Study}
	In this subsection, we conduct several ablation studies: (1) Ablation study on the two components: Stitch-Up and Co-Learning; (2) How to apply Stitch-Up augmentation? (3) Comparison against Mix-Up augmentation; (4) Effect of the sampling strategy for Stitch-Up; (5) Effect of $K$ images used in Stitch-Up augmentation; (6) Effect of the probability $p$ to apply Stitch-Up augmentation; (7) Stitch-Up brings more noisy labels? (8) Pseudo labels from Co-Learning or Self-Training? (9) Effect of different sampling priors. (10) Running time analysis. All results are reported with the mAP on VOC-MLT-Noise under the noise rate $\gamma$ of 0.5.
	\input{tables/table_ablation_stitchupmode}
	\subsubsection{Ablation analysis} 
	To further understand our proposed method, we first establish a stronger baseline with a two-branch structure based on DB~\cite{wu2020distribution}. We adopt random and balanced samplers for two branches. The loss functions remain the same as ours. This brings around 3.25\% mAP improvement upon DB~\cite{wu2020distribution}. Then, based on this strong baseline, we perform the ablation analysis on Stitch-Up augmentation and Heterogeneous Co-Learning, named Stitch-Up and Pseudo Labeling, respectively. The results can be found in Table~\ref{table:ablation}. As we can see, both Stitch-Up augmentation and Co-Learning can promote the model's performance. For Stitch-Up augmentation, we show that this augmentation receives the total mAP of 68.53\%~(+1.69\%). It suggests that training with the synthesized less noisy samples can relieve the effect of label noise. However, the augmentation can also impair the evaluation performance of head classes. And we notice that the improvement primarily comes from the medium and tail classes. We explain that Stitch-Up might affect the sampling distribution. For the component of Pseudo Labeling, the testing result shows the total mAP~(68.46\%) is significantly improved. Compared to the baseline without any additional modules, Co-Learning brings the improvement of the head classes. This can be complementary to the Stitch-Up augmentation. Finally, we obtain the best total mAP performance 69.10\%, which is better than employing Stitch-Up or Co-Learning alone. It indicates that these two mechanisms can foster learning with less noisy training samples collaboratively.
	
	\input{tables/table_ablation_mixup}
	
	\subsubsection{How to apply Stitch-Up augmentation?}
	As discussed in Section~\ref{sec:stitchup}, Stitch-Up augmentation can be implemented in various forms. We explore three regular types in deep learning based methods: input images concatenation, features concatenation and features average. Table~\ref{tab:ab_stitch_mode} shows that features average reaches the best result~(68.53\%). We notice that all three types of Stitch-Up augmentation perform better than the  baseline without Stitch-Up. Note that our Stitch-Up augmentation is easy to implement. 
	\subsubsection{Comparison against Mix-Up augmentation} Our Stitch-Up shares similarity with Mix-Up~\cite{zhang2017mixup} widely used in addressing label noise\cite{li2020dividemix,arazo2019unsupervised}. We are curious whether Stitch-Up can outperform Mix-Up in the multi-label long-tailed problem with label noise. Follow~\cite{zhang2017mixup}, Mix-Up samples the interpolation parameter from Beta distribution. For fair comparison, our Stitch-Up is applied at the image level. Table~\ref{tab:ab_mixup} shows our Stitch-Up achieves better results~(68.08\%) than Mix-Up~(63.61\%). It is seen that Mix-Up performs better on head classes and the performance on medium and tail classes drops. We hypothesize that medium and tail classes are affected more severely by multi-label noise. As we discussed in Section~\ref{sec:stitchup}, Mix-Up leads to the loss of information. Because samples on medium and tail classes are limited and often co-exist with head classes, the effect on medium and tail classes is amplified. It makes learning on medium and tail classes harder. Instead, the model focuses more on the optimization of head classes. Therefore, the model can recognize head classes better.
	\subsubsection{Effect of the sampling strategy for Stitch-Up} We investigate the effect of the sampling strategy for Stitch-Up. For random and balanced sampling, we compare our Stitch-Up with the baseline respectively. The results are reported in Table~\ref{table:stitchup_sampling}. We observe that Stitch-Up augmentation can enhance the performance no matter what sampling strategy is used. The improvement mostly benefits from the medium and tail classes while sacrificing the head classes.
	\input{tables/table_ablation_stitchup_sampling}
	\input{figures/fig_ablation_k}
	\input{figures/fig_stitchup_voc}
	\input{tables/table_ablation_samplingPL}
	\subsubsection{Effect of $K$ images used in Stitch-Up augmentation} 
	Intuitively, stitching up too many images can not make any sense and can even hurt the performance. If we stitch up the whole dataset, it is highly possible to find all the object classes in the generated new image. Such an easy training sample might force the network to learn less useful representation. We perform Stitch-Up augmentation on the two-branch baseline without Co-Learning. We conduct a series of experiments to investigate the effect of different number of images ($K$) when we employ Stitch-Up augmentation. Here, we conduct the ablation study on $K=2, 3, 4, 5$. The results with $p=1.0$ fixed are presented in Figure~\ref{fig:ab_k_exp}. We have two major observations. First, the total mAP gets worse~(67.06\%, 66.52\%) when $K$ is large~$(K=4, 5)$. This phenomenon is consistent with our conjecture. Second, we find the performance for head classes drops a lot while tail classes are less affected. In the experiments, we choose $K=2$.
	
	\subsubsection{Effect of the probability $p$ to apply Stitch-Up augmentation}
	We study the influence of the probability of this augmentation when applied to the training samples. The experiments are conducted on the two-branch Stitch-Up augmented baseline without Co-Learning. We keep $K=2$ fixed. We evaluate our Stitch-Up under $p \in \{0.0, 0.3, 0.5, 0.8, 1.0\}$. As shown in Figure~\ref{fig:ab_p_exp}, $p=1.0$ achieves the overall best performance 68.53\%, which is significantly better than the case $p=0.0$~(66.84\%) when we perform no stitch-up augmentation. Meanwhile, we observe that the stitch-up augmentation consistently receives the improvement with increasing probability $p$. This confirms that our Stitch-Up augmentation can relieve label noise by synthesizing cleaner training samples. As a result,  we set $p=1.0$ in the experiments.
	\subsubsection{Stitch-Up brings more noisy labels?} It can occur when we stitch up two images where both images contain no class objects but one of them is annotated with the noisy label of that class. Due to the feature average option, the gradients are back-propagated for both images, which can misguide the direction of optimization. However, we find that the positive effect~(reducing label noise) of Stitch-Up outweighs the negative effect~(introducing label noise) in practice. We calculate the ratio between the amount of label noise reduced and the amount of label noise introduced in one epoch on VOC-MLT-Noise. The ratio is around 2.34. It indicates the overall benefit of Stitch-Up is to reduce label noise.
	
	\subsubsection{Pseudo labels from Co-Learning or Self-Training?}
	As self-training is also a promising approach to solving the noisy label learning problem\cite{han2019deep}, can Co-Learning perform better than Self-Training? We compare the performance between Co-Learning and Self-Training in the multi-label long-tailed setting when introduced label noise. We evaluate the two strategies under different sampling distributions. Note that we conduct a hyperparameter search for Self-Training. In Table~\ref{table:study_sampling_PL}, we refer ``cross" to Co-Learning and ``self" to Self-Training. It is observed that Co-Learning achieves better total mAP performance no matter what sampling policies we use. Especially when training with random and balanced sampling simultaneously, the gap between Co-Learning~(69.75\%) and Self-Training~(67.70\%) can be expanded to 2\%. When we look at the results of each branch, Co-Learning achieves superior performance for both branches. We argue that Self-Training is prone to accumulate errors. When training with noisy labels under the long-tailed distribution, the negative impact is extremely amplified. Co-Learning leverages the inconsistency between sampling to avoid the confirmation bias with wrong noisy labels so that we can learn a robust model effectively.

\input{figures/fig_stitchup_coco}
	\input{figures/fig_vis_pl}
	
	\subsubsection{Effect of different sampling priors} We claim that the inconsistency that comes from different sampling priors can help us correct label noise. To further substantiate our hypothesis, we investigate various combinations of random sampling and balanced sampling, including random+balanced, random+random and balanced+balanced. As shown in Table~\ref{table:study_sampling_PL}, random sampling with balanced sampling can bring about 2\% performance gain while the same sampling priors for two branches can slightly improve upon baseline. It demonstrates that disagreement under different sampling priors is key to the success of Co-Learning in challenging scenarios with both label noise and long-tailed distribution.
	\input{tables/table_training_time}
	\subsubsection{Running time analysis} We compare the total inference time between DB~\cite{wu2020distribution} and our proposed method on VOC2007 clean test set in Table~\ref{table:time_compare}. We report the average running time over 5 trials on the whole test set with a batch size of 1. The experiment is conducted on a single Nvidia RTX2080Ti GPU. Our method is slower than DB due to Co-Learning. The overhead mainly comes from the two-branch ensemble architecture. It introduces an extra model forward time.
	
	\subsection{Qualitative Evaluation}
	\subsubsection{Lower noise level after Stitch-Up}
	We define the noise level as the real noise rate of the training data, different from the original noise rate $\gamma$ of the training dataset. As we discussed in Section~\ref{sec:stitchup}, Stitch-Up can synthesize cleaner training samples, resulting in a lower noise level of our training data. To verify our motivation, we visualize the change in the noise level with or without Stitch-Up during the training stage. The result on VOC-MLT-Noise and COCO-MLT-Noise is presented in Figure~\ref{fig:voc_stitchup} and Figure~\ref{fig:coco_stitchup}, respectively. It is clear to see the noise level of the training data is decreasing after we perform Stitch-Up augmentation no matter what sampling strategies we use. This demonstrates that the improvement of the performance benefits from cleaner samples introduced by Stitch-Up. We also notice that the noise level keeps steady during the whole training procedure. The noise level of the head class is reduced more compared to the medium and tail class. This is possibly due to the influence of the long-tailed distribution.
	
	\subsubsection{Visualization of pseudo labels from Co-Learning}
	We show several training images with their pseudo labels in Figure~\ref{fig:vis_pl} for an intuitive illustration of our Co-Learning. We observe that our model can assign relatively high scores to those labels that might occur in the given images and low scores to those absent labels. For example, the car class does not exist in the first example at the top. And the chair class and dining table class are missing in the annotation. Through the label correction from Co-Learning, we rectify the noisy labels by adding the chair and dining table classes and removing the car class. The generated pseudo labels are closer to the clean labels. Therefore, Co-Learning can facilitate the learning of the model with cleaner labels.
	
	\section{Conclusion}
	In this paper, we address multi-label long-tailed visual recognition with noisy labels. Training with noisy labels can hinder the development of a robust model. Considering inherent properties of multi-label classification and long-tailed learning under noisy cases, we propose a Heterogeneous Co-Learning framework equipped with a novel Stitch-Up augmentation to mitigate the impact of label noise. Through extensive experiments on two synthetic noisy datasets named VOC-MLT-Noise and COCO-MLT-Noise, we show that our method exhibits substantial results compared to various baselines.
	
	\bibliographystyle{IEEEtran}
	\bibliography{egbib}
	
\end{document}

%% file: figures/figure1.tex
\begin{figure}[t]
    \centering
    \includegraphics[width=0.88\linewidth]{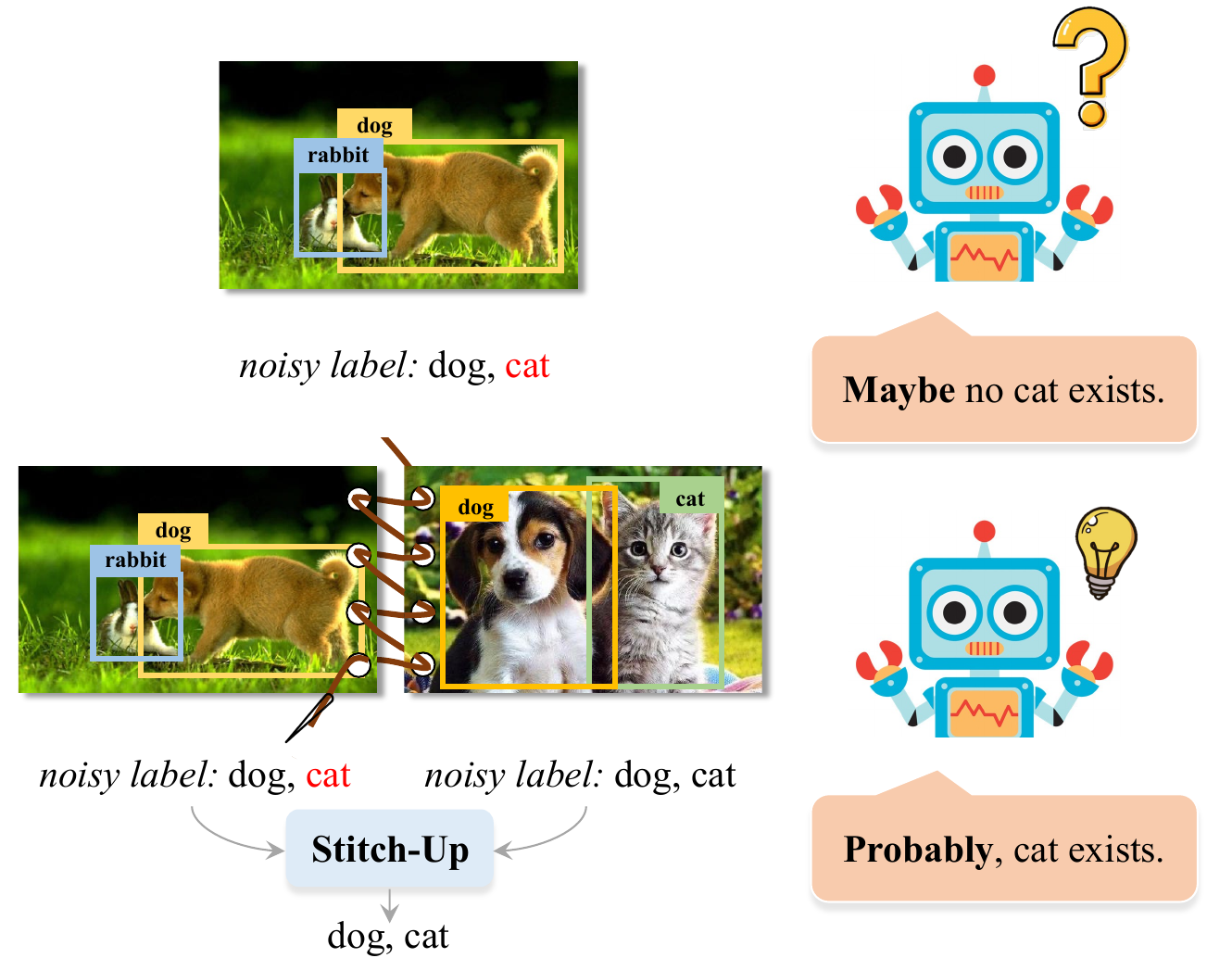}
    \caption{We show an example to illustrate that stitching up a set of images can reduce label noise. Models can be confused with the wrong \textit{noisy label} \textcolor{red}{cat} but enhance the confidence when given a set of images containing noisy label cat.}
    \label{fig:intro}
\end{figure}

%% file: figures/figure2.tex
\begin{figure*}[t]
  \centering
  \subfloat[\label{fig:framework_colearning}]{
  \includegraphics[width=0.83\linewidth]{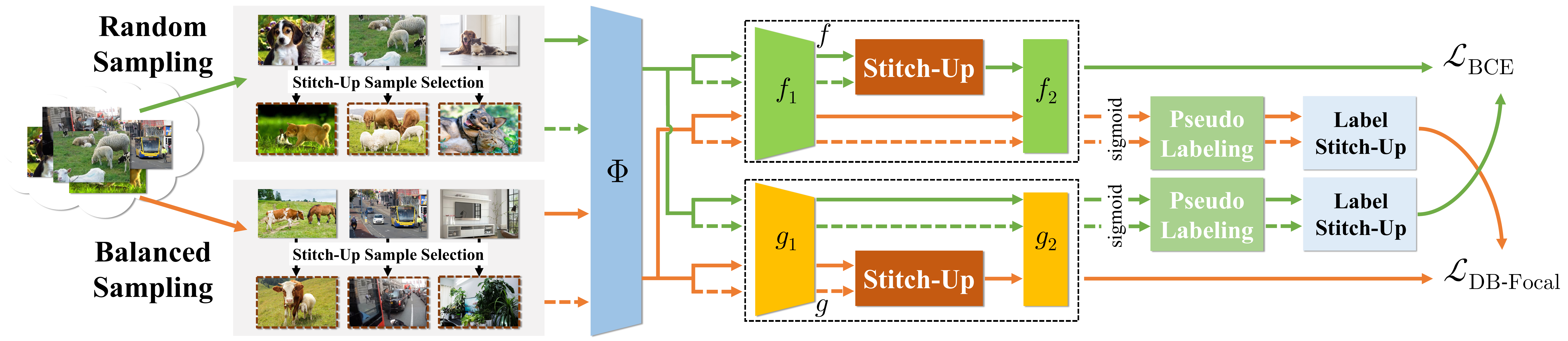}}
  
  \subfloat[\label{fig:framework_stitchup}]{
  \includegraphics[width=0.84\linewidth]{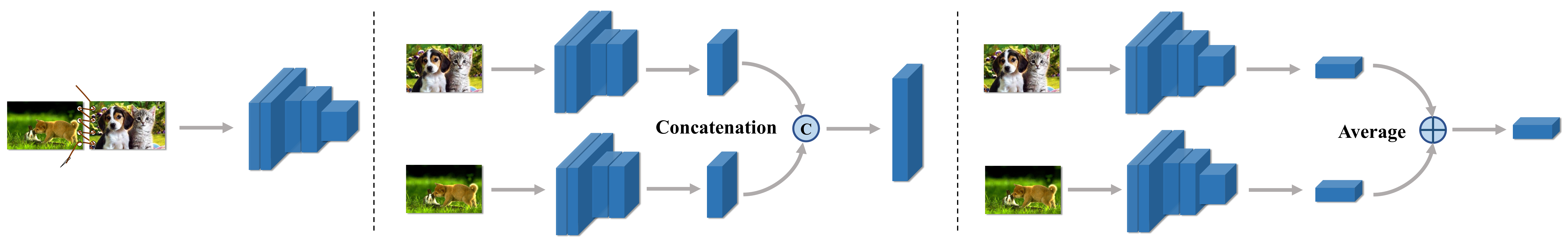}}
  
  \caption{We present the overall Heterogeneous Co-Learning framework equipped with Stitch-Up augmentation in \ref{fig:framework_colearning}. \ref{fig:framework_stitchup} shows three regular types of Stitch-Up augmentation~(from left to right): input images concatenation, features concatenation and features average~(\textbf{default}).}
  \label{fig:framework}
\end{figure*}

%% file: algo/algorithm.tex
\begin{algorithm}
\small
\SetKwData{Left}{left}\SetKwData{This}{this}\SetKwData{Up}{up}
\SetKwInOut{Input}{Input}\SetKwInOut{Output}{Output}
\LinesNumbered
\Input{Noisy training dataset $\mathcal{D}_{train}$, batch size $n, m$, max iteration $\textit{MaxIters}$}
\Output{deep neural network $\Phi, f, g$}

\For {$t=0,1,...,\text{MaxIters}-1$}{
$(\mathbf{X}_1, \mathbf{\widetilde{Y}}_1) \leftarrow RandomSampleBatch (\mathcal{D}_{train}, n)$\;
$(\mathbf{X}_1', \mathbf{\widetilde{Y}}_1') \leftarrow BalancedSampleBatch (\mathcal{D}_{train}, m)$\;
{\footnotesize\tcp{Stitch-Up}}
$(\mathbf{X}_2, \mathbf{\widetilde{Y}}_2) \leftarrow Sample Selection (\mathbf{X}_1, \mathbf{\widetilde{Y}}_1)$\;
$(\mathbf{X}_2', \mathbf{\widetilde{Y}}_2') \leftarrow Sample Selection (\mathbf{X}_1', \mathbf{\widetilde{Y}}_1')$\;
Obtain logits $\mathbf{\bar{Z}}(\mathbf{\bar{Z}}')$ by Eq~\ref{eq:stitch_feature} and Eq~\ref{eq:gen_logit}\;
{\footnotesize\tcp{Pseudo Labeling}}
Correct noisy labels $\mathbf{\widetilde{Y}}_1, \mathbf{\widetilde{Y}}_2(\mathbf{\widetilde{Y}}_1', \mathbf{\widetilde{Y}}_2')$ by Eq~\ref{eq:pseudo_label}\;
{\footnotesize\tcp{Label Stitch-Up}}
Obtain $\mathbf{\bar{Y}}(\mathbf{\bar{Y}}')$ by Eq~\ref{eq:lb_stitchup}\;
Update $\Phi, f$ by Eq~\ref{eq:bce} with $\mathbf{\bar{Z}},\mathbf{\bar{Y}}$\;
Update $\Phi, g$ by Eq~\ref{eq:db} with $\mathbf{\bar{Z}}',\mathbf{\bar{Y}}'$\;
}

\caption{Pseudo-code for our method.}
\label{alg:overall}
\end{algorithm}

%% file: tables/table_total.tex
\begin{table*}
\caption{Comparison between different methods on VOC-MLT-Noise and COCO-MLT-Noise under different noise rate settings. We report the mAP performance on the total classes. For our two-branch based methods, we also show the evaluation results for each branch. We highlight the best two results~(excluding our separate branch): \textbf{rank-1}, \underline{rank-2}. }
\centering
\resizebox{0.89\textwidth}{!}{
\begin{tabular}{l|c|c|c|c||c|c|c|c}
\toprule
Dataset &\multicolumn{4}{c||}{VOC-MLT-Noise} & \multicolumn{4}{c}{COCO-MLT-Noise} \\
\midrule
 Noise rate~$\gamma$ & \multicolumn{1}{c|}{0.3}& \multicolumn{1}{c|}{0.5} & \multicolumn{1}{c|}{0.7} & \multicolumn{1}{c||}{0.9}& \multicolumn{1}{c|}{0.3}& \multicolumn{1}{c|}{0.5} & \multicolumn{1}{c|}{0.7} & \multicolumn{1}{c}{0.9} \\ 
\midrule
ERM & 65.92 $\pm$ 0.65 & 52.21 $\pm$ 0.71 & 41.32 $\pm$ 0.83 & 17.24 $\pm$ 0.37 & 39.94 $\pm$ 0.17 & 34.64 $\pm$ 0.30 & 27.76 $\pm$ 0.28 & 19.12 $\pm$ 0.44 \\
Focal Loss~\cite{lin2017focal} & 69.19 $\pm$ 0.19 & 55.47 $\pm$ 0.67 & 44.86 $\pm$ 0.40 & 19.32 $\pm$ 0.41 & 46.90 $\pm$ 0.23 & 41.85 $\pm$ 0.14 & 33.71 $\pm$ 0.37 & 22.00 $\pm$ 0.28 \\
RS~\cite{shen2016relay}  & 71.84 $\pm$ 0.74 & 58.54 $\pm$ 0.70 & 52.51 $\pm$ 1.16 & \underline{29.09 $\pm$ 0.28} & 49.10 $\pm$ 0.24 & 44.66 $\pm$ 0.20 & 37.18 $\pm$ 0.53 & 25.44 $\pm$ 0.48 \\
RS-Focal & 71.29 $\pm$ 0.34 & 57.90 $\pm$ 0.50 & 53.02 $\pm$ 1.17 & 25.74 $\pm$ 0.65 & 51.05 $\pm$ 0.13 & 46.16 $\pm$ 0.19 & 38.64 $\pm$ 0.37 & 25.12 $\pm$ 0.36 \\
LDAM~\cite{cao2019learning} & 67.09 $\pm$ 0.51 & 53.12 $\pm$ 0.66	& 40.42 $\pm$ 0.65 & 17.58 $\pm$ 0.24 & 40.28 $\pm$ 0.29 & 35.05 $\pm$ 0.58 & 27.19 $\pm$ 0.22 & 19.16 $\pm$ 0.40 \\
BBN~\cite{zhou2020bbn} & 58.23 $\pm$ 0.68 & 47.39 $\pm$ 1.57 & 40.24 $\pm$ 1.44 & 21.99 $\pm$ 0.57 & 32.43 $\pm$ 0.57 & 29.63 $\pm$ 0.08 & 26.36 $\pm$ 0.26 & 20.54 $\pm$ 0.10 \\
ML-GCN~\cite{chen2019multi} & 67.76 $\pm$ 0.44 & 54.98 $\pm$ 0.71 & 49.07 $\pm$ 0.57 & 25.19 $\pm$ 0.63 & 48.02 $\pm$ 0.18 & 44.31 $\pm$ 0.27 & 37.28 $\pm$ 0.14 & 25.56 $\pm$ 0.28 \\
DivideMix~\cite{li2020dividemix} &  68.25 $\pm$ 0.54 & 59.30 $\pm$ 0.81 & 48.98 $\pm$ 1.02 & 27.52 $\pm$ 0.68 & 37.26 $\pm$ 0.32 &  34.52 $\pm$ 0.39 & 31.77 $\pm$ 0.46 & 23.28 $\pm$ 0.36 \\
DB~\cite{wu2020distribution} & \underline{73.75} $\pm$ 0.16	& \underline{63.59 $\pm$ 0.50} & 53.89 $\pm$ 0.93	& 27.41 $\pm$ 0.23 & 51.95 $\pm$ 0.20 & 48.03 $\pm$ 0.11 & \underline{42.65 $\pm$ 0.17} & \underline{29.88 $\pm$ 0.32} \\
DB-Focal~\cite{wu2020distribution} & 72.87 $\pm$ 0.27 & 61.48 $\pm$ 0.47 & \underline{55.02 $\pm$ 1.57} & 27.20 $\pm$ 1.14 & \underline{52.44 $\pm$ 0.23} & \underline{48.43 $\pm$ 0.34} & 42.61 $\pm$ 0.23 & 28.68 $\pm$ 0.28 \\
\midrule
Ours & \textbf{76.48 $\pm$ 0.36} & \textbf{69.10 $\pm$ 0.23} & \textbf{62.29 $\pm$ 0.94} & \textbf{34.41 $\pm$ 0.51} & \textbf{54.14 $\pm$ 0.16} & \textbf{50.42 $\pm$ 0.04} & \textbf{45.47 $\pm$ 0.30} & \textbf{33.10 $\pm$ 0.23}\\
\rowcolor{Gray}
Ours-random & 64.14 $\pm$ 1.14 & 56.16 $\pm$ 1.25 & 50.09 $\pm$ 0.56 & 27.58 $\pm$ 0.87 & 43.55 $\pm$ 0.38 & 40.77 $\pm$ 0.26 & 33.69 $\pm$ 0.60 & 26.79 $\pm$ 0.55\\
\rowcolor{Gray}
Ours-balance & 74.96 $\pm$ 0.28 & 67.57 $\pm$ 0.33 & 59.47 $\pm$ 0.90 & 32.16 $\pm$ 0.50 & 52.54 $\pm$ 0.18 & 48.38 $\pm$ 0.06 & 43.50 $\pm$ 0.26 & 31.31 $\pm$ 0.35\\

\bottomrule
\end{tabular}}

\label{table:total}
\end{table*}

%% file: tables/table_voc.tex
\begin{table*}
\caption{Comparison between different methods on VOC-MLT-Noise under different noise rate settings. We report the mAP performance on three subsets including head, medium, and tail. For our two-branch based methods, we also show the evaluation results for each branch. We highlight the best two results~(excluding our separate branch): \textbf{rank-1}, \underline{rank-2}. }
\centering
\setlength{\tabcolsep}{2pt}
\resizebox{0.9\textwidth}{!}{
\begin{tabular}{l|ccc|ccc|ccc|ccc}
\toprule
\multicolumn{13}{c}{VOC-MLT-Noise} \\
\midrule
 Noise rate~$\gamma$ & \multicolumn{3}{c|}{0.3}& \multicolumn{3}{c|}{0.5} & \multicolumn{3}{c|}{0.7} & \multicolumn{3}{c}{0.9}\\ 
\midrule
 Methods & head & medium & tail & head & medium & tail & head & medium & tail & head & medium & tail \\
\midrule
ERM & 59.85 $\pm$ 0.43 & 74.66 $\pm$ 0.54 & 63.92 $\pm$ 1.66 & 51.52 $\pm$ 0.84 & 64.29 $\pm$ 0.98 & 43.68 $\pm$ 1.24 & 34.89 $\pm$ 0.62 & 49.69 $\pm$ 0.86 & 39.85 $\pm$ 1.26 & 21.77 $\pm$ 0.49 & 16.92 $\pm$ 1.13 & 14.07 $\pm$ 1.00\\
Focal Loss~\cite{lin2017focal} & 59.34 $\pm$ 0.75 & 75.72 $\pm$ 0.54 & 71.68 $\pm$ 0.38 & 50.10 $\pm$ 0.92 & 64.51 $\pm$ 0.66 & 52.72 $\pm$ 1.38 & 34.46 $\pm$ 0.37& 50.49 $\pm$ 0.87 & 48.43 $\pm$ 1.14 & 21.48 $\pm$ 0.33 & 17.25 $\pm$ 0.24 & 19.25 $\pm$ 0.85\\
RS~\cite{shen2016relay}  & 63.50 $\pm$ 0.64 & 79.51 $\pm$ 0.44 & 72.35 $\pm$ 1.47 & 50.55 $\pm$ 0.37 & 72.12 $\pm$ 1.41 & 54.36 $\pm$ 1.49 & 37.65 $\pm$ 0.97 & \underline{67.31 $\pm$ 1.19} & 52.56 $\pm$ 3.23 & 31.12 $\pm$ 1.31 & \underline{37.54 $\pm$ 0.68} & 21.23 $\pm$ 0.91\\
RS-Focal & 61.07 $\pm$ 0.63 & 78.65 $\pm$ 0.48 & 73.43 $\pm$ 0.89 & 48.99 $\pm$ 1.16 & 72.33 $\pm$ 0.88 & 53.75 $\pm$ 1.70 & 37.84 $\pm$ 0.75 & 65.13 $\pm$ 1.65 & 55.33 $\pm$ 1.89 & 28.99 $\pm$ 0.34 & 29.39 $\pm$ 1.14 & 20.56 $\pm$ 1.65\\
LDAM~\cite{cao2019learning} & 59.74 $\pm$ 0.36 & 74.21 $\pm$ 0.33 & 67.25 $\pm$ 1.58 & 51.11 $\pm$ 0.44 & 65.97 $\pm$ 0.95 & 45.00 $\pm$ 1.95 & 34.09 $\pm$ 0.66 & 49.78 $\pm$ 0.67 & 38.14 $\pm$ 1.87 & 21.59 $\pm$ 0.28 & 16.31 $\pm$ 0.71 & 15.53 $\pm$ 0.43\\
BBN~\cite{zhou2020bbn} & \underline{67.03 $\pm$ 0.98} & 69.74 $\pm$ 0.55 & 42.99 $\pm$ 2.36 & \underline{61.99 $\pm$ 0.54} & 62.57 $\pm$ 2.74 & 25.05 $\pm$ 2.55 & \underline{50.09 $\pm$ 0.91} & 54.40 $\pm$ 1.42 & 22.23 $\pm$ 2.72 & \underline{36.60 $\pm$ 0.97} & 23.18 $\pm$ 0.74 & 10.14 $\pm$ 1.08\\
ML-GCN~\cite{chen2019multi} & 64.75 $\pm$ 1.38 &74.09 $\pm$ 0.40 & 65.27 $\pm$ 0.57 & 60.84 $\pm$ 1.42 & 69.11 $\pm$ 1.01 & 39.98 $\pm$ 1.71 & 48.03 $\pm$ 1.66 & 58.04 $\pm$ 2.47 & 43.13 $\pm$ 1.04 & 33.80 $\pm$ 0.57 & 28.64 $\pm$ 1.54 & 16.14 $\pm$ 2.03\\
DivideMix~\cite{li2020dividemix} & 60.30 $\pm$ 1.83 & 76.84 $\pm$ 0.71 & 67.77 $\pm$ 1.08 & 58.20 $\pm$ 0.70 & 69.11 $\pm$ 1.22 & 52.78 $\pm$ 0.87 & 36.55 $\pm$ 1.37 & 66.52 $\pm$ 1.73 & 45.14 $\pm$ 2.41 & 36.14 $\pm$ 1.63 & 31.51 $\pm$ 2.14 & 18.06 $\pm$ 0.97 \\
DB~\cite{wu2020distribution} & 65.55 $\pm$ 0.52 & \underline{80.22 $\pm$ 0.35} & 75.04 $\pm$ 0.53 & 57.90 $\pm$ 1.00 & \underline{73.76 $\pm$ 0.51} & \underline{60.22 $\pm$ 1.12} & 41.09 $\pm$ 1.03 & 62.11 $\pm$ 1.80 & \underline{57.34 $\pm$ 1.25} & 33.72 $\pm$ 1.78 & 26.35 $\pm$ 1.53 & \underline{23.49 $\pm$ 2.00}\\
DB-Focal~\cite{wu2020distribution} & 64.23 $\pm$ 0.50 & 78.09 $\pm$ 0.31	& \underline{75.43 $\pm$ 0.26} & 56.91 $\pm$ 0.36 & 73.20 $\pm$ 0.56 & 56.11 $\pm$ 1.07 & 43.63 $\pm$ 1.03 & 65.46 $\pm$ 1.31 & 55.73 $\pm$ 2.92 & 31.93 $\pm$ 0.65 & 30.57 $\pm$ 3.55 & 21.13 $\pm$ 1.08\\
\midrule
Ours & \textbf{67.85 $\pm$ 0.95} & \textbf{80.87 $\pm$ 0.54} & \textbf{79.67 $\pm$ 0.76} & \textbf{66.40 $\pm$ 0.38} & \textbf{77.71 $\pm$ 0.38} & \textbf{64.67 $\pm$ 0.83} & \textbf{57.40 $\pm$ 1.15} & \textbf{72.40 $\pm$ 0.87} & \textbf{58.37 $\pm$ 2.23} & \textbf{41.71 $\pm$ 0.51} & \textbf{38.36 $\pm$ 1.00} & \textbf{25.98 $\pm$ 1.03} \\
\rowcolor{Gray}
Ours-random & 69.23 $\pm$ 1.02 & 78.38 $\pm$ 0.55 & 49.65 $\pm$ 2.49 & 65.39 $\pm$ 0.84 & 73.76 $\pm$ 0.34 & 36.04 $\pm$ 2.59 & 57.74 $\pm$ 2.15 & 68.99 $\pm$ 0.75 & 30.18 $\pm$ 1.93 & 39.29 $\pm$ 1.63 & 34.03 $\pm$ 2.62 & 13.96 $\pm$ 1.33 \\
\rowcolor{Gray}
Ours-balance & 64.71 $\pm$ 0.69 & 79.15 $\pm$ 0.58 & 79.51 $\pm$ 0.76 & 64.99 $\pm$ 0.55 & 75.67 $\pm$ 0.93 & 63.42 $\pm$ 0.75 & 53.48 $\pm$ 0.61 & 67.47 $\pm$ 1.01 & 57.95 $\pm$ 2.02 & 39.52 $\pm$ 1.10 & 33.78 $\pm$ 0.87 & 25.42 $\pm$ 0.87 \\

\bottomrule
\end{tabular}}

\label{table:VOC}
\end{table*}

%% file: tables/table_coco.tex
\begin{table*}
\caption{Comparison between different methods on COCO-MLT-Noise under different noise rate settings. We report the mAP performance on three subsets including head, medium, and tail. For our two-branch based methods, we also show the evaluation results for each branch. We highlight the best two results~(excluding our separate branch): \textbf{rank-1}, \underline{rank-2}. }
\centering
\setlength{\tabcolsep}{2pt}
\resizebox{0.9\textwidth}{!}{
\begin{tabular}{l|ccc|ccc|ccc|ccc}
\toprule
\multicolumn{13}{c}{COCO-MLT-Noise} \\
\midrule
 Noise rate~$\gamma$ & \multicolumn{3}{c|}{0.3}& \multicolumn{3}{c|}{0.5} & \multicolumn{3}{c|}{0.7} & \multicolumn{3}{c}{0.9}\\ 
\midrule
 Methods & head & medium & tail & head & medium & tail & head & medium & tail & head & medium & tail \\
\midrule
ERM & 54.59 $\pm$ 0.24 & 43.73 $\pm$ 0.31 & 19.34 $\pm$ 0.48 & 48.70 $\pm$ 0.35 & 38.59 $\pm$ 0.66 & 14.47 $\pm$ 0.54 & 39.28 $\pm$ 0.31 & 31.87 $\pm$ 0.68 & 10.13 $\pm$ 0.61 & 29.87 $\pm$ 0.37 & 22.01 $\pm$ 1.02 & 3.87 $\pm$ 0.29\\
Focal Loss~\cite{lin2017focal} & 54.56 $\pm$ 0.33 & 47.17 $\pm$ 0.08 & 38.26 $\pm$ 0.48 & 49.11 $\pm$ 0.24 & 43.34 $\pm$ 0.17 & 32.13 $\pm$ 0.54 & 39.03 $\pm$ 0.32 & 36.99 $\pm$ 0.34 & 23.83 $\pm$ 0.83 & 28.98 $\pm$ 0.27 & 25.02 $\pm$ 0.64 & 10.68 $\pm$ 0.56\\
RS~\cite{shen2016relay}  & 54.90 $\pm$ 0.21 & 48.04 $\pm$ 0.44 & 44.13 $\pm$ 0.45 & 50.44 $\pm$ 0.40 & 45.52 $\pm$ 0.39 & 37.32 $\pm$ 0.17 & 42.46 $\pm$ 0.27 & 38.62 $\pm$ 0.68 & 29.66 $\pm$ 1.00 & 34.29 $\pm$ 0.55 & 27.52 $\pm$ 1.10 & 13.24 $\pm$ 0.74\\
RS-Focal & 55.87 $\pm$ 0.11 & 49.75 $\pm$ 0.29 & 47.44 $\pm$ 0.42 & 50.76 $\pm$ 0.37 & 46.17 $\pm$ 0.47 & 41.16 $\pm$ 0.24 & 42.29 $\pm$ 0.23 & 39.59 $\pm$ 0.46 & 33.49 $\pm$ 0.72 & 32.69 $\pm$ 0.45 & 27.35 $\pm$ 0.32 & 14.12 $\pm$ 1.06\\
LDAM~\cite{cao2019learning} & 54.52 $\pm$ 0.17 & 44.32 $\pm$ 0.34 & 19.82 $\pm$ 0.68 & 48.87 $\pm$ 0.39 & 38.85 $\pm$ 0.88 & 15.33 $\pm$ 1.00 & 38.60 $\pm$ 0.44 & 31.41 $\pm$ 0.76 & 9.55 $\pm$ 0.73 & 29.32 $\pm$ 0.56 & 22.58 $\pm$ 0.77 & 3.88 $\pm$ 0.26\\
BBN~\cite{zhou2020bbn} & 50.32 $\pm$ 0.39 & 32.80 $\pm$ 0.72 & 12.59 $\pm$ 1.73 & 49.21 $\pm$ 0.21 & 28.93 $\pm$ 0.36 & 9.30 $\pm$ 0.57 & 45.21 $\pm$ 0.45 & 25.53 $\pm$ 0.62 & 6.99 $\pm$ 0.89 & 38.86 $\pm$ 0.54 & 17.72 $\pm$ 0.42 & 4.21 $\pm$ 0.41\\
ML-GCN~\cite{chen2019multi} & 55.76 $\pm$ 0.16 & 45.12 $\pm$ 0.14 & 43.25 $\pm$ 0.73 & 52.83 $\pm$ 0.17 & 42.32 $\pm$ 0.30 & 37.57 $\pm$ 0.86 & 46.38 $\pm$ 0.32 & 36.04 $\pm$ 0.35 & 28.96 $\pm$ 0.56 & 38.08 $\pm$ 0.39 & 25.45 $\pm$ 0.31 & 12.13 $\pm$ 0.81\\
DivideMix~\cite{li2020dividemix} & 44.30 $\pm$ 0.37 & 36.37 $\pm$ 0.35 & 30.75 $\pm$ 0.64 & 41.02 $\pm$ 0.31 & 34.01 $\pm$ 0.60 & 28.11 $\pm$ 0.52 & 39.06 $\pm$ 0.35 & 30.48 $\pm$ 0.46 & 25.49 $\pm$ 0.78 & 33.37 $\pm$ 0.34 & 22.52 $\pm$ 0.51 & 13.29 $\pm$ 0.31  \\
DB~\cite{wu2020distribution} & 56.22 $\pm$ 0.28 & 50.34 $\pm$ 0.23 & 49.34 $\pm$ 0.41 & 52.96 $\pm$ 0.18 & 47.78 $\pm$ 0.27 & 42.99 $\pm$ 0.27 & 48.23 $\pm$ 0.40 & \underline{42.98 $\pm$ 0.13} & 36.19 $\pm$ 0.41 & \underline{40.83 $\pm$ 0.31} & \underline{31.65 $\pm$ 0.46} & \underline{15.81 $\pm$ 0.66}\\
DB-Focal~\cite{wu2020distribution} & \underline{56.74 $\pm$ 0.20} & \underline{50.57 $\pm$ 0.22} & \underline{50.11 $\pm$ 0.45}	& \underline{53.51 $\pm$ 0.43} & \underline{48.03 $\pm$ 0.31} & \underline{43.43 $\pm$ 0.48} & \underline{48.38 $\pm$ 0.38} & 41.94 $\pm$ 0.31 & \underline{37.20 $\pm$ 0.50} & 39.98 $\pm$ 0.37	& 29.53 $\pm$ 0.65 & 15.37 $\pm$ 0.45\\
\midrule
Ours & \textbf{59.80 $\pm$ 0.05} & \textbf{51.53 $\pm$ 0.26} & \textbf{51.27 $\pm$ 0.30} & \textbf{56.78 $\pm$ 0.16} & \textbf{49.11 $\pm$ 0.26} & \textbf{45.17 $\pm$ 0.28} & \textbf{51.42 $\pm$ 0.32} & \textbf{45.48 $\pm$ 0.39} & \textbf{38.99 $\pm$ 0.69} & \textbf{44.52 $\pm$ 0.26} & \textbf{34.78 $\pm$ 0.47} & \textbf{18.65 $\pm$ 0.43} \\
\rowcolor{Gray}
Ours-random & 59.74 $\pm$ 0.07 & 46.64 $\pm$ 0.33 & 22.14 $\pm$ 1.01 & 56.70 $\pm$ 0.26 & 44.08 $\pm$ 0.34 & 19.37 $\pm$ 0.73 & 52.01 $\pm$ 0.49 & 37.37 $\pm$ 0.61 & 9.23 $\pm$ 0.99 & 44.12 $\pm$ 0.58 & 28.54 $\pm$ 0.92 & 5.85 $\pm$ 0.53 \\
\rowcolor{Gray}
Ours-balance & 56.61 $\pm$ 0.22 & 50.25 $\pm$ 0.19 & 51.00 $\pm$ 0.37 & 52.90 $\pm$ 0.13 & 47.50 $\pm$ 0.23 & 44.60 $\pm$ 0.28 & 47.34 $\pm$ 0.29 & 43.68 $\pm$ 0.28 & 39.11 $\pm$ 0.77 & 41.31 $\pm$ 0.33 & 32.74 $\pm$ 0.62 & 18.69 $\pm$ 0.50 \\

\bottomrule
\end{tabular}}

\label{table:COCO}
\end{table*}

%% file: tables/table_ablation.tex
\begin{table}
\caption{Ablation study on Stitch-Up and Co-Learning. We report the mAP performance on VOC-MLT-Noise. PL denotes Pseudo Labeling.}
\centering
\small
\setlength{\tabcolsep}{9pt}
\resizebox{0.78\linewidth}{!}{
\begin{tabular}{cc|cccc}
    \toprule
    \multicolumn{6}{c}{VOC-MLT-Noise} \\
    \midrule
    Stitch-Up & PL & total & head & medium & tail \\
    \midrule
    & & 66.84 & 65.33 & 77.02 & 60.33 \\
    \cmark & & 68.53 & 63.33 & 79.05 & 64.55 \\
    & \cmark & 68.46 & 68.77 & 76.98 & 61.83 \\
    \cmark & \cmark & 69.10 & 66.40 & 77.71 & 64.67 \\
    \bottomrule
    \end{tabular}}
    \label{table:ablation}
\end{table}

%% file: tables/table_ablation_stitchupmode.tex
\begin{table}
\caption{Different implementations for Stitch-Up augmentation. We report the mAP performance on VOC-MLT-Noise. concat.~: concatenation, \xmark~: no Stitch-Up. }
\centering
\small
\setlength{\tabcolsep}{6.8pt}
\resizebox{0.78\linewidth}{!}{
\begin{tabular}{c|cccc}
    \toprule
    \multicolumn{5}{c}{VOC-MLT-Noise} \\
    \midrule
    Stitch-Up mode & total & head & medium & tail \\
    \midrule
    average before $f_3$ & 68.53 & 63.33 & 79.05 & 64.55\\
    concat. before GAP & 67.73 & 63.65 & 78.48 & 62.73\\
    concat. input images & 68.08 & 64.27 & 78.44 & 63.17\\
    \xmark & 66.84 & 65.33 & 77.02 & 60.33 \\
    \bottomrule
    \end{tabular}}
    \label{tab:ab_stitch_mode}
\end{table}

%% file: tables/table_ablation_mixup.tex
\begin{table}
\caption{Comparison between Stitch-Up and Mix-Up. We report the mAP performance on VOC-MLT-Noise. \xmark~: no augmentation.}
\centering
\small
\setlength{\tabcolsep}{9pt}
\resizebox{0.78\linewidth}{!}{
\begin{tabular}{c|cccc}
    \toprule
    \multicolumn{5}{c}{VOC-MLT-Noise} \\
    \midrule
    Augmentation & total & head & medium & tail \\
    \midrule
    Stitch-Up & 68.08 & 64.27 & 78.44 & 63.17\\
    Mix-Up & 63.61 & 67.27 & 75.06 & 52.26 \\
    \xmark & 66.84 & 65.33 & 77.02 & 60.33 \\
    \bottomrule
    \end{tabular}}
    \label{tab:ab_mixup}
    \vspace{-0.3cm}
\end{table}

%% file: tables/table_ablation_stitchup_sampling.tex
\begin{table}
\caption{Effect of the sampling strategy for Stitch-Up. We report the mAP performance on VOC-MLT-Noise.}
\centering
\small
\setlength{\tabcolsep}{6pt}
\resizebox{0.78\linewidth}{!}{
\begin{tabular}{c|c|cccc}
    \toprule
    \multicolumn{6}{c}{VOC-MLT-Noise} \\
    \midrule
    Sampling & Stitch-Up & total & head & medium & tail \\
    \midrule
    \multirow{2}{*}{Random} & & 53.35 & 64.19 & 72.62 & 30.77 \\
    & \cmark & 58.06 & 63.88 & 76.11 & 40.15 \\
    \midrule
    \multirow{2}{*}{Balanced}&  & 64.03 & 60.34 & 73.75 & 59.53 \\
    & \cmark & 65.86 & 58.14 & 76.98 & 63.30 \\
    \bottomrule
    \end{tabular}}
    \label{table:stitchup_sampling}
\end{table}

%% file: figures/fig_ablation_k.tex
\begin{figure}[ht]
  \subfloat[\label{fig:ab_k_exp}]{
  \includegraphics[width=0.45\linewidth]{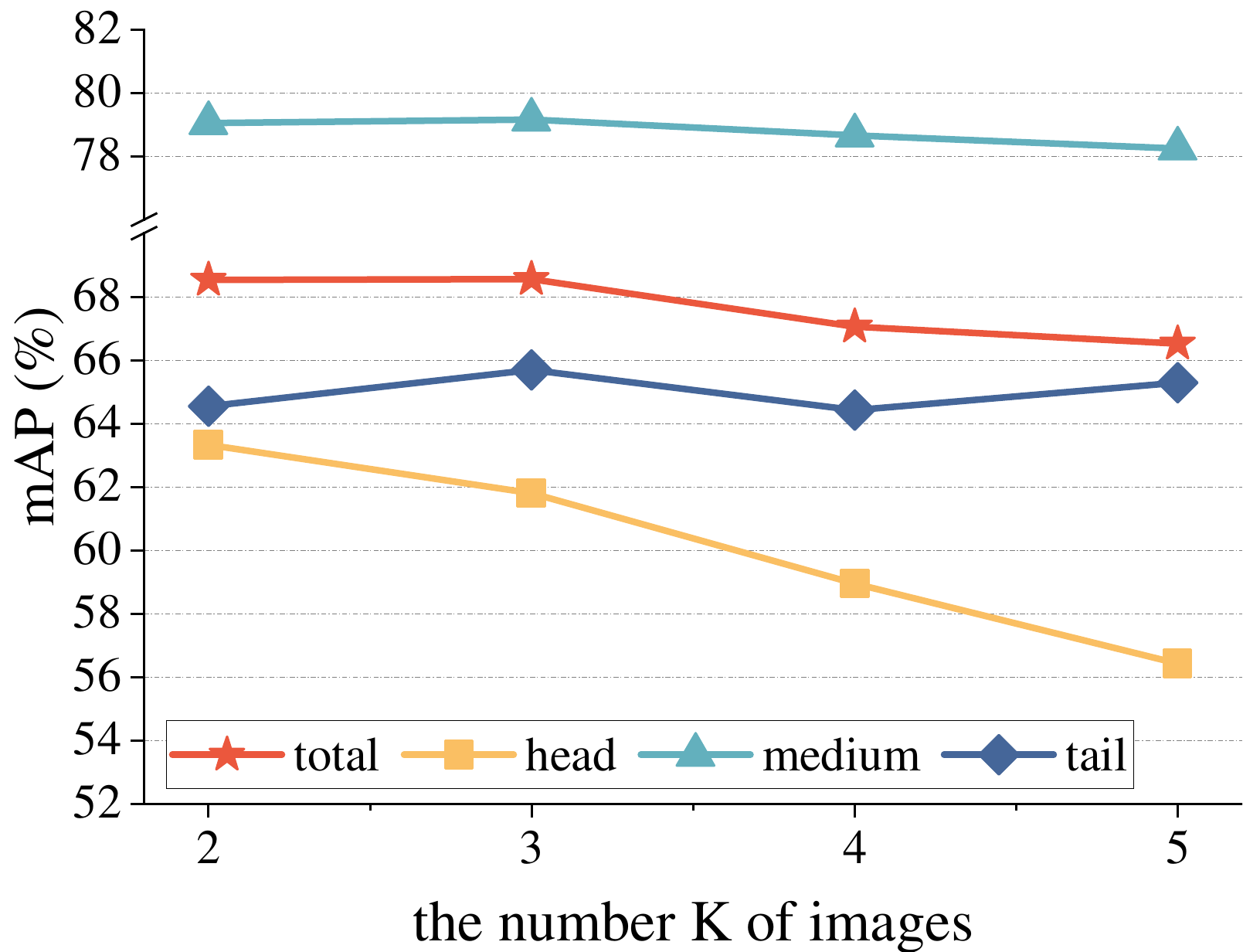}}
  \subfloat[\label{fig:ab_p_exp}]{
  \includegraphics[width=0.45\linewidth]{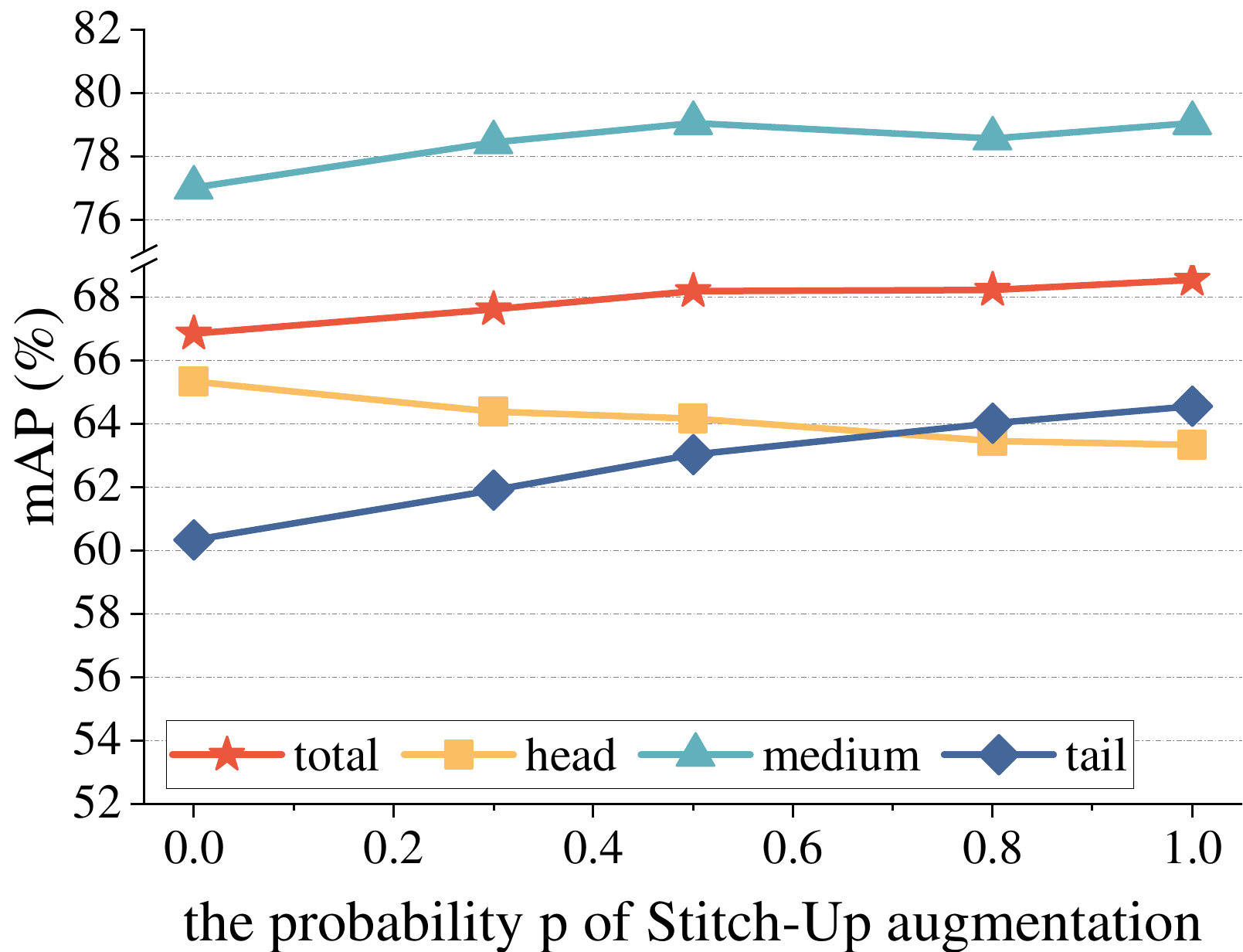}}
  \caption{\ref{fig:ab_k_exp}: Ablation study on the effect of the number of images for Stitch-Up augmentation. \ref{fig:ab_p_exp}: Ablation study on the effect of the probability of Stitch-Up augmentation applied to training samples. We report the mAP performance on VOC-MLT-Noise under the noise rate of 0.5.}
  \label{fig:voc_ab_kp}
\end{figure}

%% file: figures/fig_stitchup_voc.tex
\begin{figure*}[t]
  \centering
  \subfloat[\label{fig:voc_all}]{
  \includegraphics[width=0.27\textwidth]{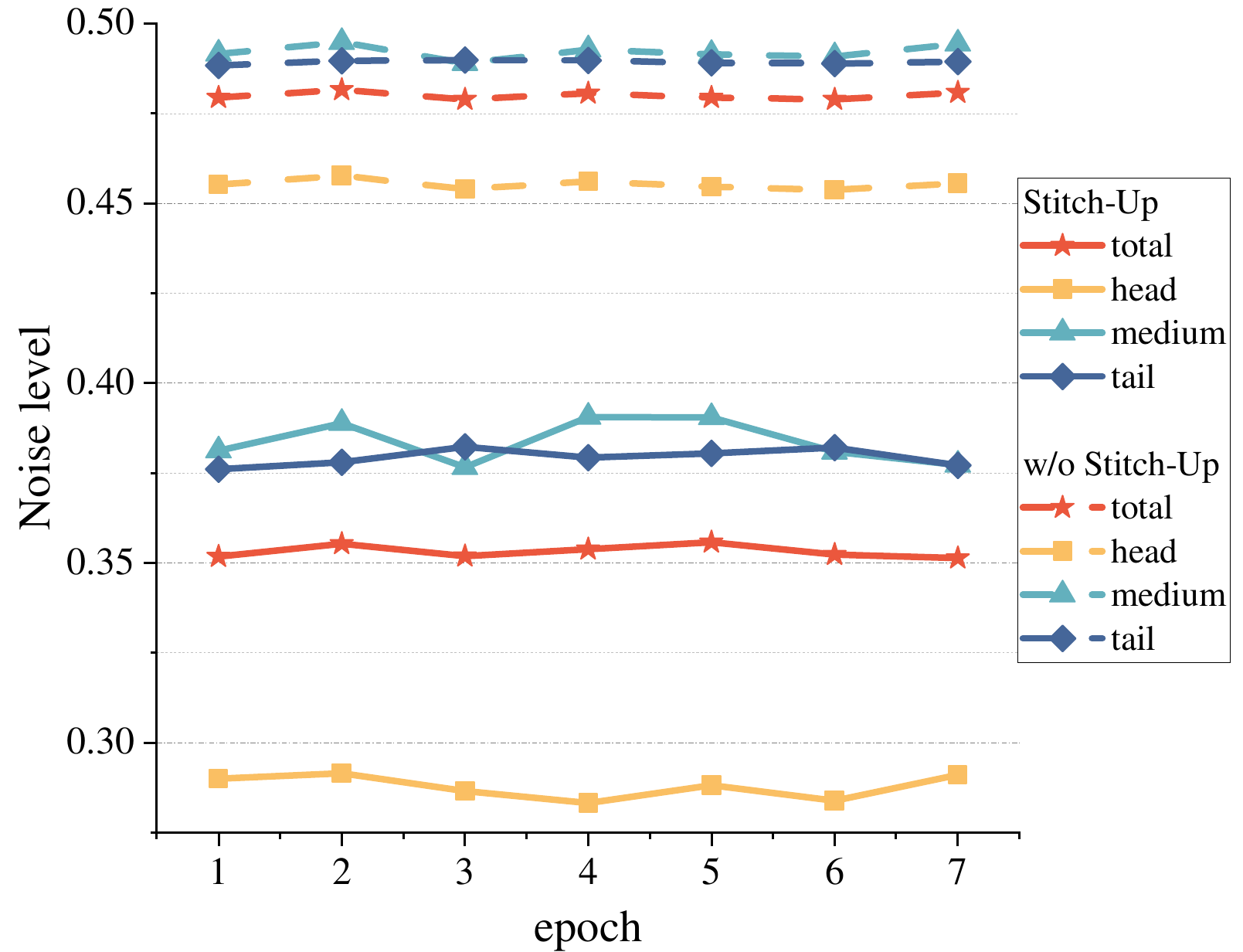}}
  \subfloat[\label{fig:voc_random}]{
  \includegraphics[width=0.27\textwidth]{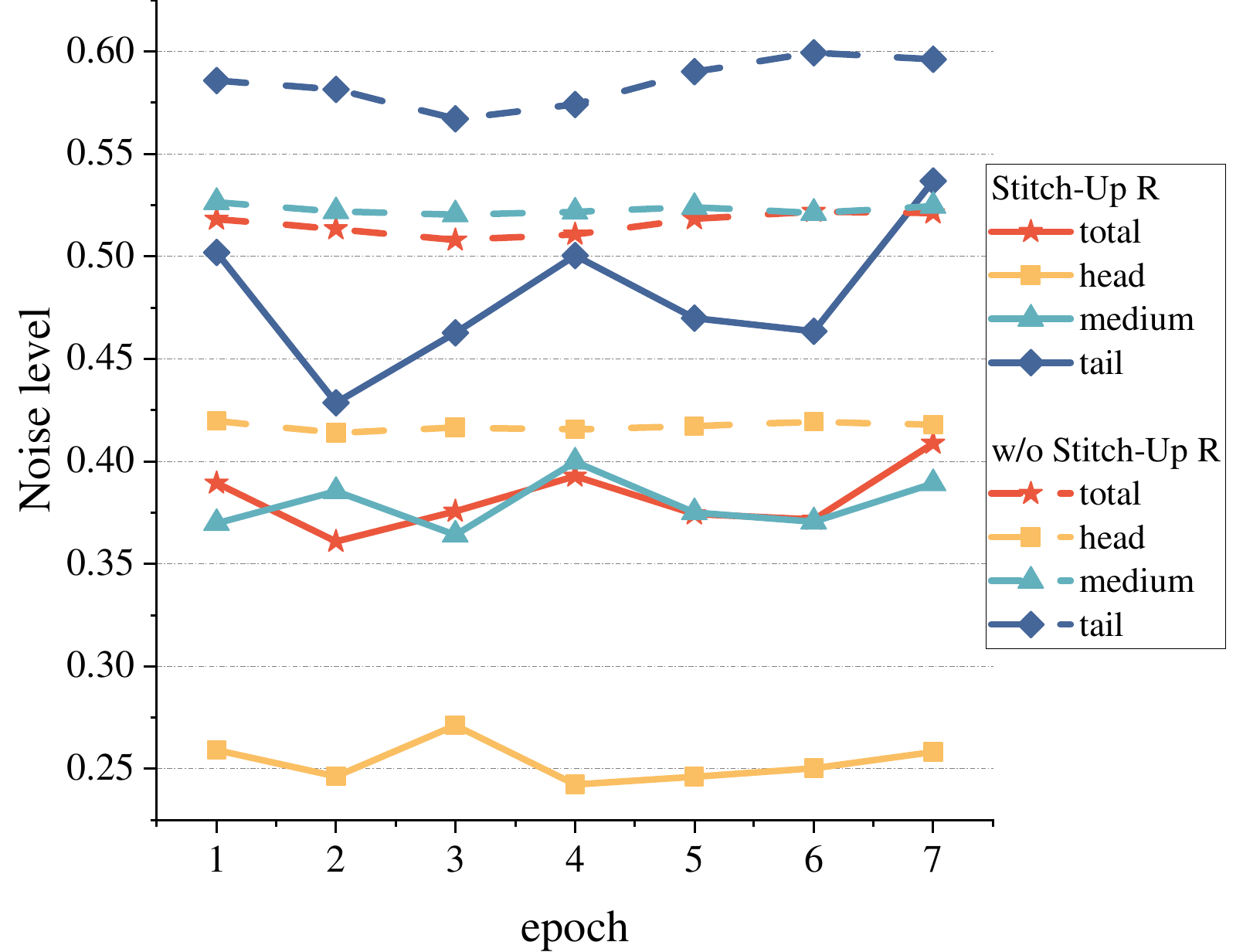}}
  \subfloat[\label{fig:voc_balance}]{
  \includegraphics[width=0.27\textwidth]{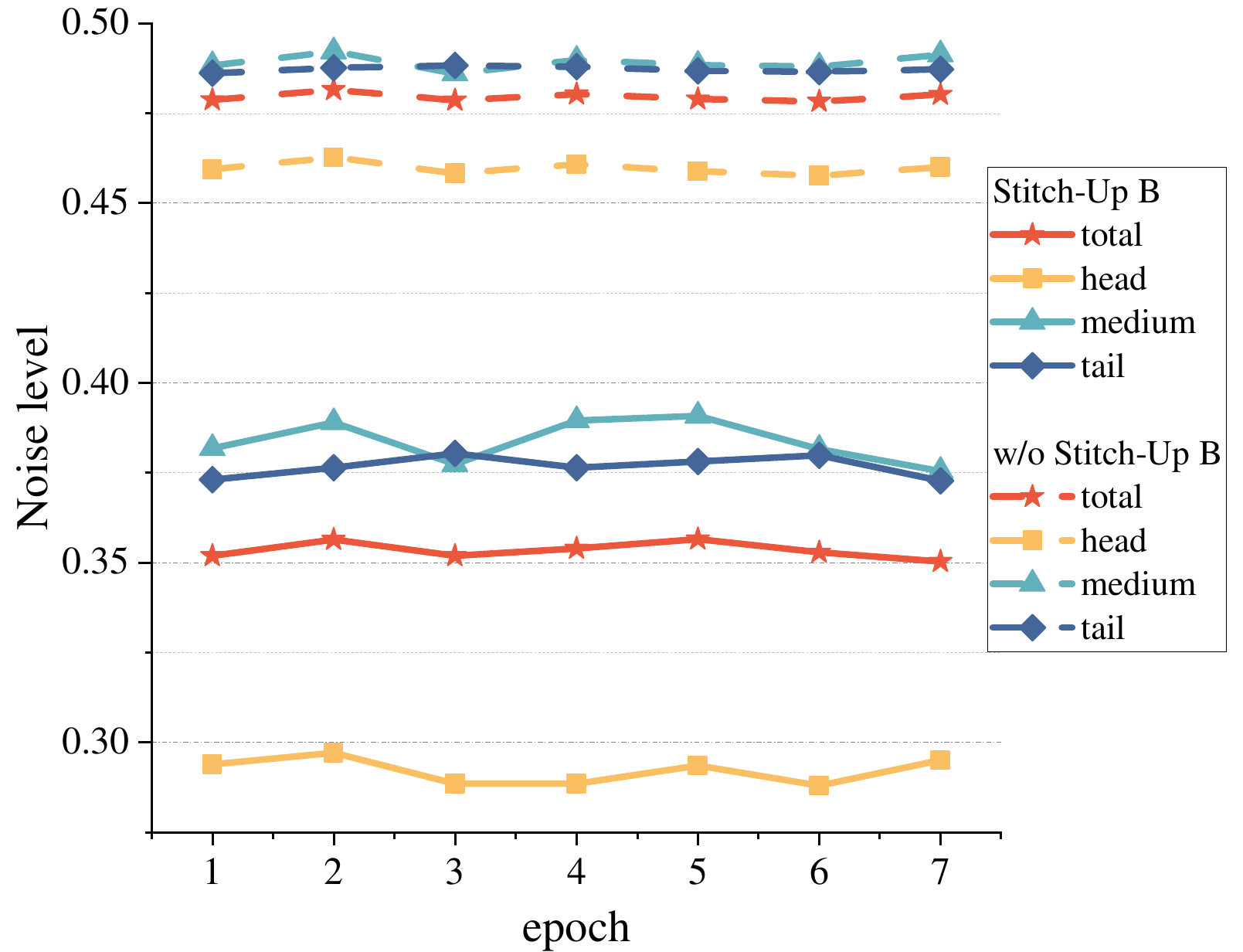}}
  \caption{We visualize the change of the noise level on VOC-MLT-Noise under the noise rate of 0.5 during the training. \ref{fig:voc_all} presents the overall two branch. \ref{fig:voc_random} presents the change in the random branch. \ref{fig:voc_balance} presents the change in the balance branch. R denotes the random branch. B denotes the balance branch.}
  \label{fig:voc_stitchup}
\end{figure*}

%% file: tables/table_ablation_samplingPL.tex
\begin{table}
\caption{Effects of Pseudo Label and Sampling priors in Heterogeneous Co-Learning framework. We report the mAP on VOC-MLT-Noise. We also show the evaluation results for each branch. S denotes the combinations of sampling, including R+B~(random sampling+balanced sampling), R+R~(random sampling+random sampling), B+B~(balanced sampling+balanced sampling). PL denotes where pseudo labels are from, including \xmark~(no pseudo labels), self~(pseudo labels from Self-Training), cross~(pseudo-labels from Co-Learning). We highlight the best two results in the ensemble branch: \textbf{rank-1},  \underline{rank-2}.}
\centering
\small
\resizebox{0.84\linewidth}{!}{
\begin{tabular}{c|c|c|cccc}

\toprule
\multicolumn{7}{c}{VOC-MLT-Noise} \\

\midrule
 S & PL & test branch & total & head & medium & tail \\

\midrule
 \multirow{9}{*}{R+B} & \multirow{3}{*}{\xmark} & ensemble & 67.35 & 64.28 & \underline{78.51} & \underline{61.29}\\
 ~ & ~ & random & 54.77 & 63.49 & 77.05 & 31.51\\
 ~ & ~ & balance & 64.55 & 59.49 & 75.48 & 60.16\\

 \cmidrule{2-7}
  ~ & \multirow{3}{*}{self} & ensemble & \underline{67.70} & 66.53 & 78.31 & 60.63\\
  ~ & ~ & random & 52.83 & 66.25 & 75.30 & 25.92\\
  ~ & ~ & balance & 64.18 & 59.22 & 75.62 & 59.33\\

 \cmidrule{2-7}
  ~ & \multirow{3}{*}{cross} & ensemble & \textbf{69.75} & \textbf{67.55} & \textbf{78.95} & \textbf{64.50}\\
  ~ & ~ & random & 54.32 & 66.72 & 76.16 & 28.65\\
  ~ & ~ & balance & 68.18 & 65.03 & 76.66 & 64.19\\
  
 \midrule
  
 \multirow{9}{*}{R+R} & \multirow{3}{*}{\xmark} & ensemble & 63.26 & 63.05 & 76.34 & 53.62\\
 ~ & ~ & random & 55.30 & 63.58 & 75.41 & 34.01\\
 ~ & ~ & balance & 59.75 & 57.93 & 72.02 & 51.92\\

 \cmidrule{2-7}
  ~ & \multirow{3}{*}{self} & ensemble & 63.98 & 65.53 & 76.14 & 53.70\\
  ~ & ~ & random & 54.60 & 65.36 & 74.91 & 31.30\\
  ~ & ~ & balance & 59.75 & 57.93 & 72.02 & 51.92\\

 \cmidrule{2-7}
  ~ & \multirow{3}{*}{cross} & ensemble & 64.54 & \underline{67.42} & 75.67 & 54.03\\
  ~ & ~ & random & 54.71 & 66.13 & 74.80 & 31.07\\
  ~ & ~ & balance & 62.71 & 65.62 & 71.46 & 53.98\\
  
\midrule
 \multirow{9}{*}{B+B} & \multirow{3}{*}{\xmark} & ensemble & 60.55 & 54.67 & 70.26 & 57.69\\
 ~ & ~ & random & 53.05 & 53.26 & 66.43 & 42.85\\
 ~ & ~ & balance & 54.46 & 45.03 & 61.32 & 56.38\\

 \cmidrule{2-7}
  ~ & \multirow{3}{*}{self} & ensemble & 60.68 & 55.06 & 70.43 & 57.58\\
  ~ & ~ & random & 52.14 & 52.10 & 65.97 & 41.79\\
  ~ & ~ & balance & 54.52 & 45.13 & 61.44 & 56.38\\

 \cmidrule{2-7}
  ~ & \multirow{3}{*}{cross} & ensemble & 60.92 & 55.99 & 70.39 & 57.51\\
  ~ & ~ & random & 53.08 & 53.29 & 66.46 & 42.89\\
  ~ & ~ & balance & 56.13 & 50.65 & 61.57 & 56.16\\

\bottomrule
\end{tabular}}
\label{table:study_sampling_PL}
\end{table}

%% file: figures/fig_stitchup_coco.tex
\begin{figure*}[ht]
  \centering
  \subfloat[\label{fig:coco_all}]{
  \includegraphics[width=0.27\textwidth]{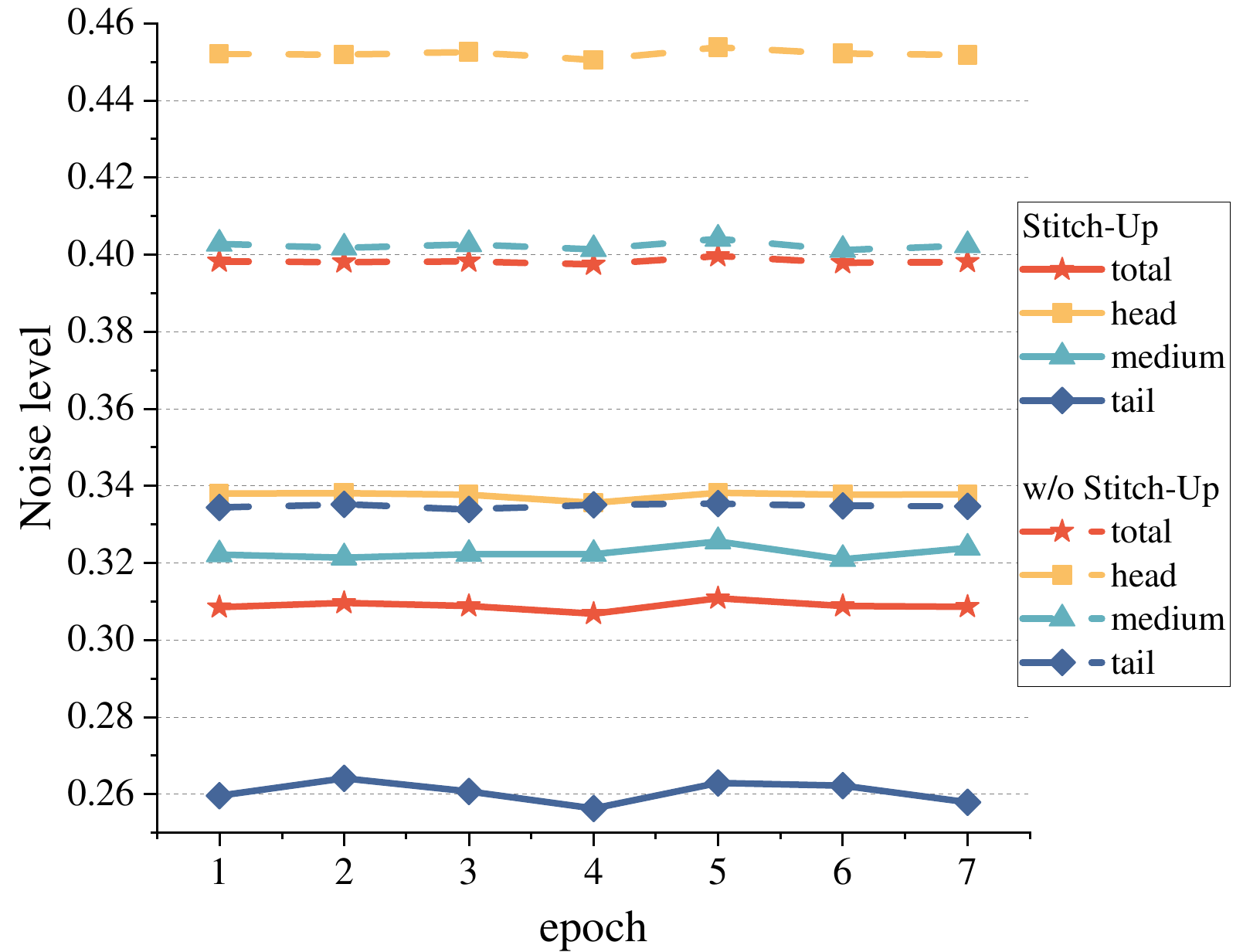}}
  \subfloat[\label{fig:coco_random}]{
  \includegraphics[width=0.27\textwidth]{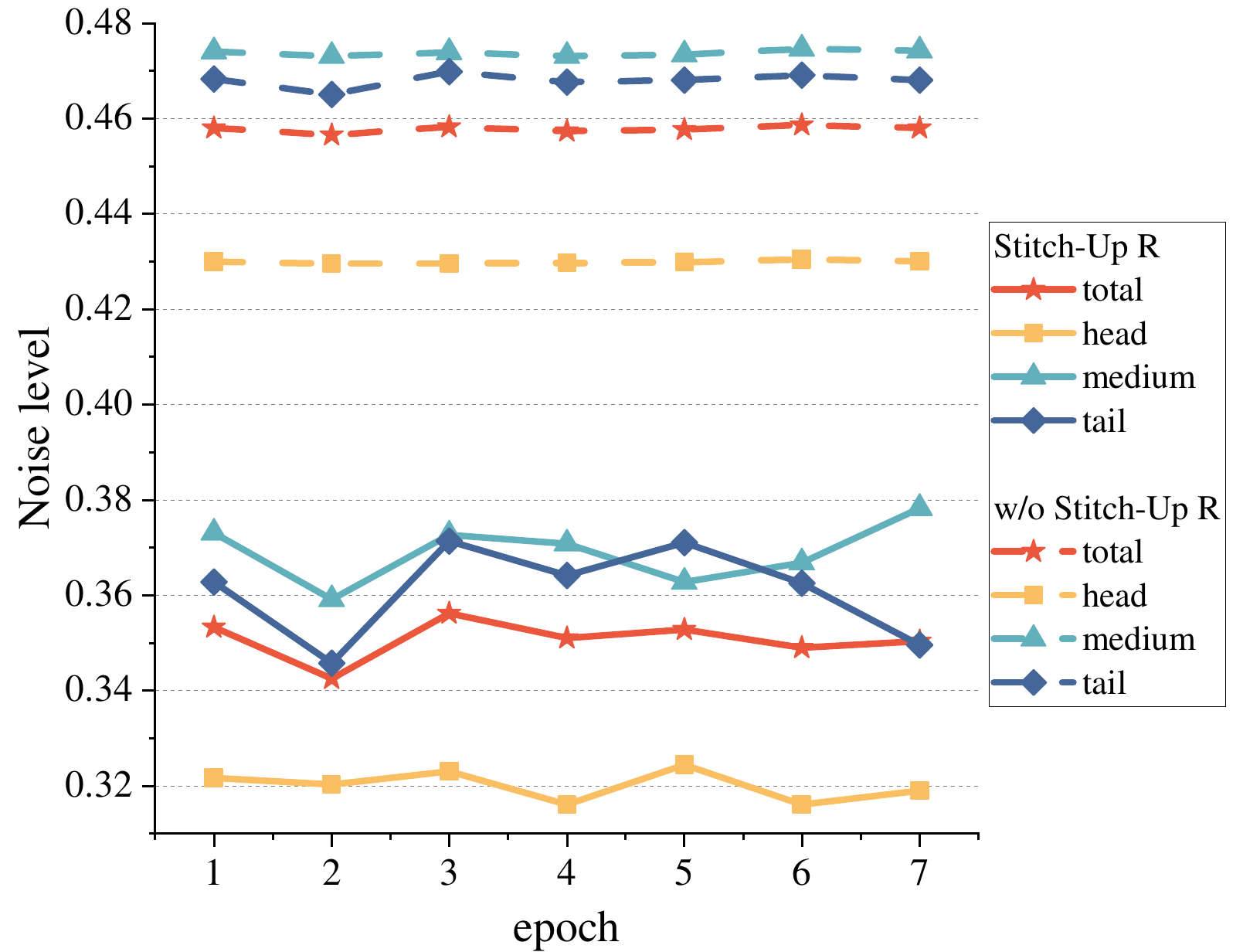}}
  \subfloat[\label{fig:coco_balance}]{
  \includegraphics[width=0.27\textwidth]{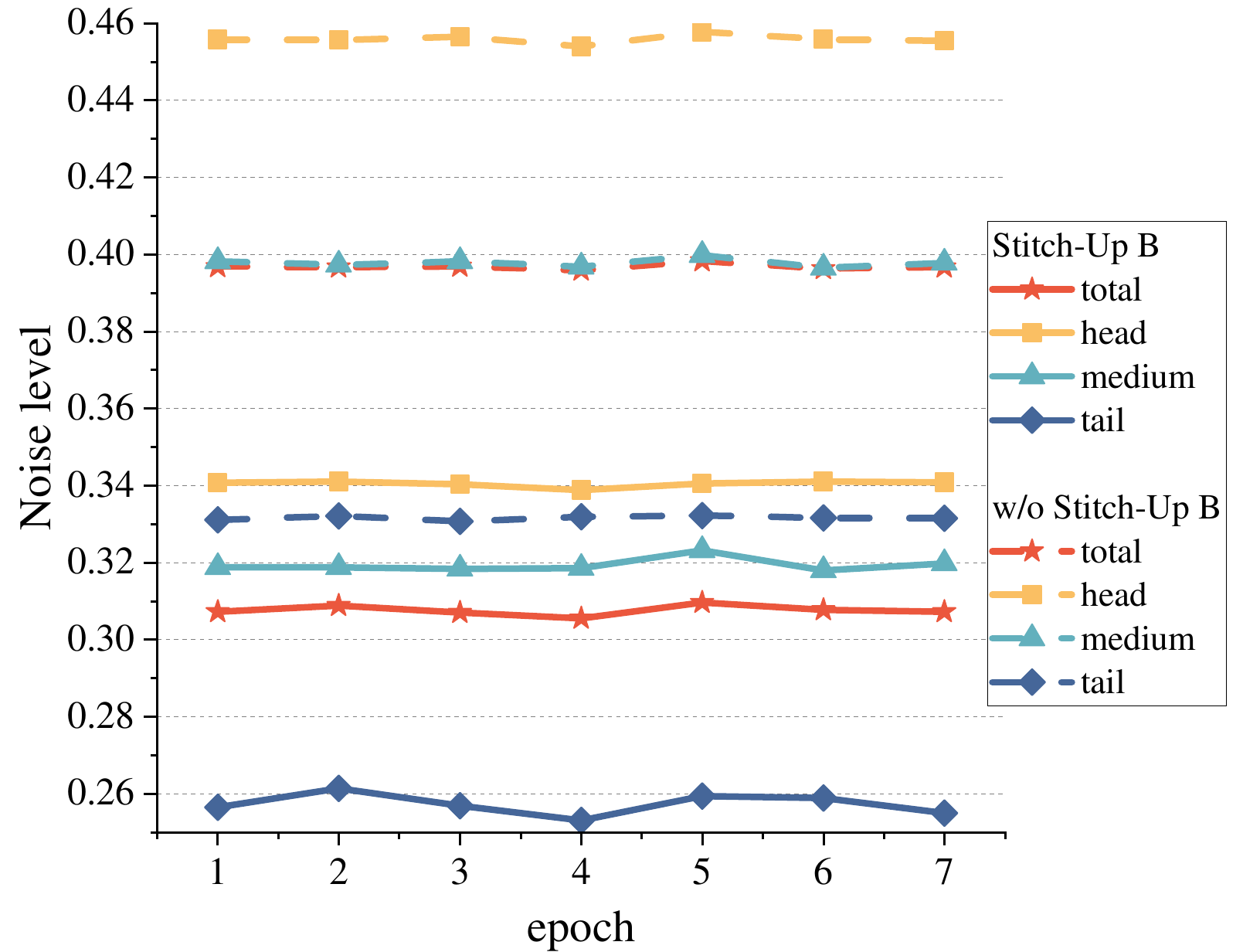}}
  \caption{We visualize the change of the noise level on COCO-MLT-Noise under the noise rate of 0.5 during the training. \ref{fig:coco_all} presents the overall two branch. \ref{fig:coco_random} presents the change in the random branch. \ref{fig:coco_balance} presents the change in the balance branch. R denotes the random branch. B denotes the balance branch.}
  \label{fig:coco_stitchup}
\end{figure*}

%% file: figures/fig_vis_pl.tex
\begin{figure*}
    \centering
    \includegraphics[width=0.8\linewidth]{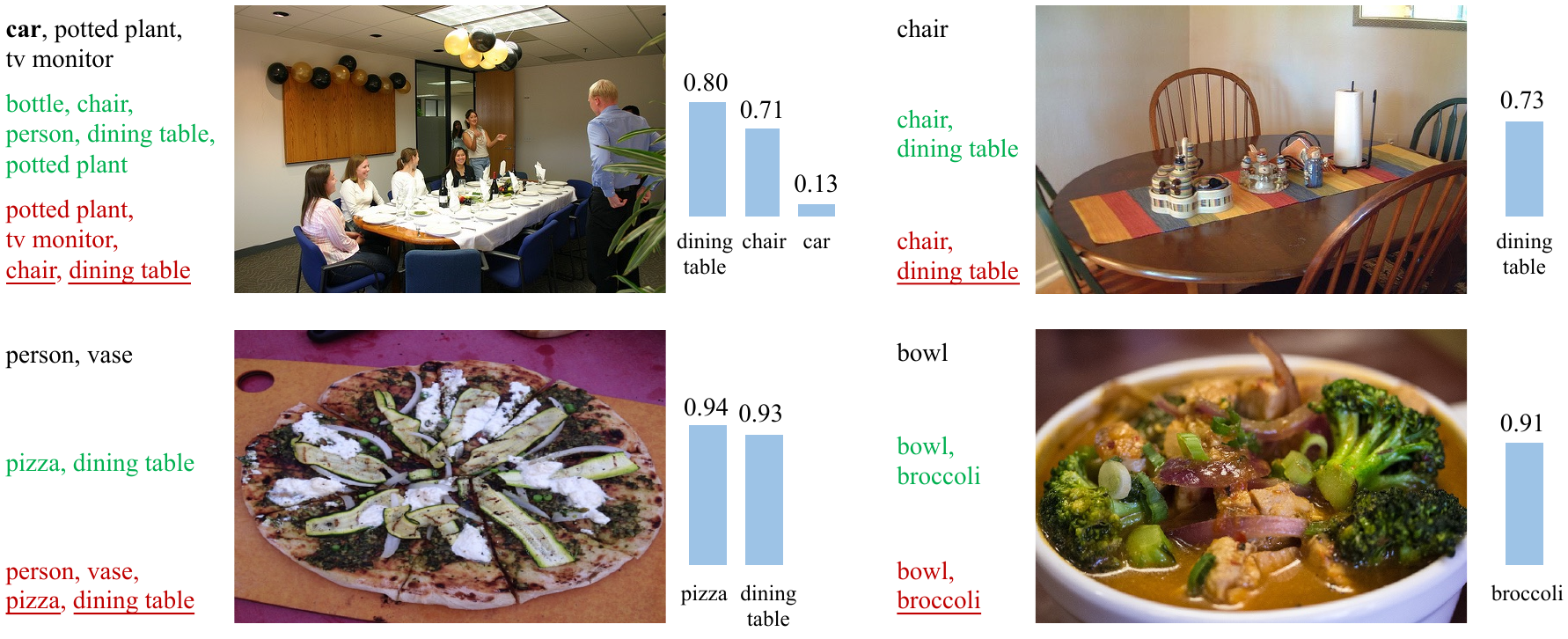}
    \caption{Several training examples on VOC-MLT-Noise~(top row) and COCO-MLT-Noise~(bottom row). We show the noisy labels, \textcolor[RGB]{0,176,80}{clean labels} as well as \textcolor[RGB]{192,0,0}{pseudo labels} given by our Co-Learning. The prediction from the model is presented on the right of the image. Here, \underline{label} denotes the missing label corrected by the model and \textbf{label} denotes the wrong label that should be discarded based on the low prediction.}
    \label{fig:vis_pl}
\end{figure*}

%% file: tables/table_training_time.tex
\begin{table}[ht]
\caption
		{
			Comparison of total inference time (seconds) on VOC2007 clean test set.
		}
	\centering
	\begin{tabular}{c|c|c} 
		\toprule	 	
			 Method & DB~\cite{wu2020distribution} & Ours \\
			\midrule
			Running time~(s) & 77.06 & 116.37  \\     
		\bottomrule
\end{tabular}
\label{table:time_compare}
\end{table}			